\documentclass{article}

\usepackage{xcolor}

\definecolor{uclablue}{RGB}{39, 116, 174}
\definecolor{bigaired}{RGB}{156, 0, 0}

\usepackage[colorlinks,citecolor=uclablue, linkcolor=bigaired]{hyperref}

    \usepackage[final]{neurips_2024}

\usepackage[utf8]{inputenc} % allow utf-8 input
\usepackage[T1]{fontenc}    % use 8-bit T1 fonts
\usepackage{url}            % simple URL typesetting
\usepackage{booktabs}       % professional-quality tables
\usepackage{amsfonts}       % blackboard math symbols
\usepackage{nicefrac}       % compact symbols for 1/2, etc.
\usepackage{microtype}      % microtypography
\usepackage{xcolor}         % colors
\usepackage{subcaption}
\usepackage{tabularx}

\usepackage{algorithm}
\usepackage{algorithmic}

\usepackage{graphicx}
\graphicspath{figure/}
\usepackage{amsmath,amsfonts,bm}
\usepackage{amsthm}
 \usepackage{amssymb}
\usepackage{mathtools}
\usepackage{enumitem}
\usepackage{wrapfig}
\usepackage{multirow}
\usepackage{multicol}
\usepackage{makecell}
\usepackage{extarrows}
\usepackage{xspace}
\usepackage[capitalize,noabbrev]{cleveref}

\usepackage{bbm}
\usepackage{xcolor,colortbl}
\usepackage{diagbox}
\usepackage{verbatim}
\usepackage{framed}
\usepackage{tcolorbox}
\usepackage{booktabs}
\usepackage{titletoc}
\usepackage{appendix}
\usepackage{pifont}
\usepackage{acronym}
\usepackage{scalerel}
\usepackage{amssymb} 

\makeatletter
\DeclareRobustCommand\onedot{\futurelet\@let@token\@onedot}
\def\@onedot{\ifx\@let@token.\else.\null\fi\xspace}
\def\eg{\emph{e.g}\onedot} 

\def\ie{\emph{i.e}\onedot}

\def\etc{\emph{etc}\onedot} 
\def\vs{\emph{vs}\onedot}

\makeatother

\makeatletter
\renewcommand{\paragraph}{%
  \@startsection{paragraph}{4}%
  {\z@}{0ex \@plus 0ex \@minus 0ex}{-1em}%
  {\normalfont\normalsize\bfseries}%
}
\makeatother

\usepackage[misc]{ifsym}

\newcommand{\ours}{\texttt{CREAM}\xspace}
\newcommand{\oursfull}{\textbf{C}ontinuity-\textbf{R}elativity ind\textbf{E}xing with g\textbf{A}ussian \textbf{M}iddle\xspace}
\everypar{\looseness=-1}

\theoremstyle{plain}
\newtheorem{theorem}{Theorem}[section]

\newtheorem{lemma}[theorem]{Lemma}

\theoremstyle{definition}
\newtheorem{definition}[theorem]{Definition}

\theoremstyle{remark}

\DeclareMathOperator*{\argmax}{arg\,max}

\title{An Efficient Recipe for Long Context Extension via Middle-Focused Positional Encoding}

\makeatletter
\def\thanks#1{\protected@xdef\@thanks{\@thanks
        \protect\footnotetext{#1}}}
\makeatother

\author{
Tong Wu \\ \texttt{wutong1@bigai.ai} \And Yanpeng Zhao \\ \texttt{zhaoyanpeng@bigai.ai} \And Zilong Zheng$^{\,\textrm{\Letter}}$\\ \texttt{zlzheng@bigai.ai}
 \AND
\normalfont State Key Laboratory of General Artificial Intelligence, BIGAI, Beijing, China \\
$^{\,\textrm{\Letter}}$ Corresponding author.
}
\begin{document}

\maketitle

\begin{abstract}

Recently, many methods have been developed to extend the context length of pre-trained large language models (LLMs), but they often require fine-tuning at the target length ($\gg4K$) and struggle to effectively utilize information from the middle part of the context. To address these issues, we propose \oursfull (\ours), which interpolates positional encodings by manipulating position indices. Apart from being simple, \ours is training-efficient: it only requires fine-tuning at the pre-trained context window (\eg, Llama 2-4K) and can extend LLMs to a much longer target context length (\eg, 256K).
To ensure that the model focuses more on the information in the middle, we introduce a truncated Gaussian to encourage sampling from the middle part of the context during fine-tuning, thus alleviating the ``Lost-in-the-Middle'' problem faced by long-context LLMs. Experimental results show that \ours successfully extends LLMs to the target length for both Base and Chat versions of \texttt{Llama2-7B} with ``Never Miss A Beat''. Our code is publicly available at \url{https://github.com/bigai-nlco/cream}. \looseness=-1
\end{abstract}

\begin{figure}[ht!]
    \centering
    \begin{subfigure}{0.49\textwidth}
        \centering
        \includegraphics[width=\linewidth, height=1.1in]{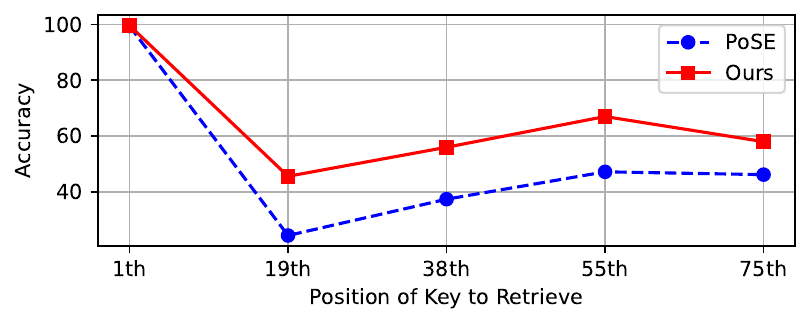}
        \caption{Linear Interpolation}
    \end{subfigure}
    \hfill
    \begin{subfigure}{0.49\textwidth}
        \centering
        \includegraphics[width=\linewidth, height=1.1in]{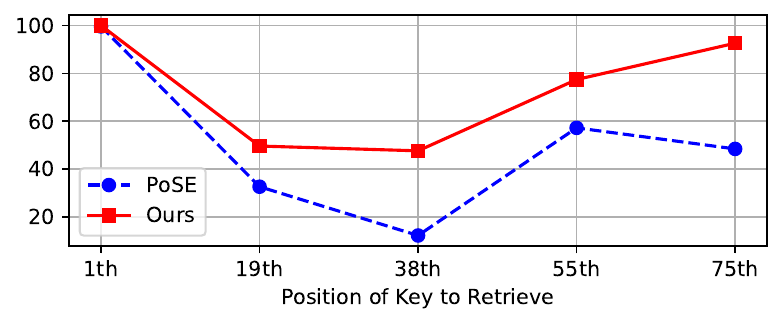}
        \caption{YaRN Interpolation}
    \end{subfigure}
    \caption{Results of applying different position interpolation methods to the ``Lost-in-the-Middle'' task on \ours and PoSE~\citep{PoSE}. We can see that \ours outperforms PoSE~\citep{PoSE} at every position, with a particularly improvement in the middle.}
    \label{fig:lost_in_the_middle}
\end{figure}

\section{Introduction}

Transformer-based Large Language Models (LLMs) are typically pre-trained with a fixed context window size, \eg, 4K tokens in \citet{llama2}. 
However, many downstream applications, 
including in-context learning~\citep{huang2023boosting,li2023loogle} and LLM agents~\citep{qian2023communicative,langsuite2023} necessitate the processing of significantly longer contexts, \eg, up to 256K tokens.
Recent works have proposed promising approaches that efficiently extend the context window of pre-trained LLMs by interpolating
Positional Encodings~(PEs)~\citep{PI,NTK,YaRN,ABF,EABF} with a short period of fine-tuning. Unlike other techniques such as efficient transformer~\citep{tworkowski2024focused,munkhdalai2024leave} and memory augmentation~\citep{tan2024lloco}, 
PE-based methods do not necessitate alterations to the model's architecture or the incorporation of supplementary modules. Consequently, PE-based methods offer the advantages of straightforward implementation and rapid adaptation, making them a practical solution for extending the operational range of LLMs in tasks involving larger context windows.

Despite the simplicity and effectiveness, existing PE-based methods exhibit two significant limitations.
\textbf{First,} prior approaches, such as positional interpolation~\citep{PI}, still require fine-tuning on the target context window size, which imposes a substantial \textit{computational overhead}~\citep{PoSE}.  
\textbf{Secondly,} though some PE methods have demonstrated potential in handling extremely long sequences, as evidenced by low sliding window perplexity scores, their performance deteriorates notably in ``in-the-middle'' scenarios~\citep{Lost_in_the_Middle}.  Specifically, when the model is required to accurately retrieve and process content located in the middle of an extended context, there is a marked drop in performance on the extended window size (\cref{fig:lost_in_the_middle} and \cref{fig:longchat_lines}). 

These observations and insights underscore a fundamental question: \textit{Can we extend the context window size of pre-trained LLMs \textbf{efficiently} while simultaneously optimizing their \textbf{effectiveness} in processing "in-the-middle" content?} 

To answer the above question, we propose \textbf{\ours}, namely \oursfull. \ours is a novel PE-based fine-tuning recipe that shows both efficiency in fine-tuning and effectiveness in enhanced middle content understanding. Our key insights lie in manipulating the positional indices of long target sequences to produce shorter ones within the pre-trained context window size (\cref{fig:frame}).

In \cref{sec:preliminaries}, we summarize two crucial ingredients of effective positional indices: continuity that produces densely connected positional indices and relativity that reveals the long-range dependencies between fragments. \ours is a recipe designed with the best of both worlds by introducing two indexing strategies for continuity and relativity, respectively (\cref{sec:method}).  Besides, to alleviate the ``Lost-in-the-Middle'' challenge, we introduce truncated Gaussian distribution for middle segment sampling, enabling the LLM to prioritize the information in the middle positions, even when performing positional interpolation within the pre-trained context window size. 

In \cref{sec:experiments}, we conduct comprehensive experiments to demonstrate the efficiency and effectiveness of \ours. We continually pre-trained on \texttt{Llama 2-7B} with \ours for a short period and extend the context window size from 4K to up to 256K. Furthermore, we instruction tuning on \texttt{Llama 2-7B-Chat} with \ours for 100 steps and obtain promising results. We highlight our empirical advantages as:
\begin{enumerate}[leftmargin=20pt, noitemsep, topsep=0pt]
\item \ours can not only fine-tune within the pre-training context window size, but also alleviate the issue of the model easily getting lost in the middle. \eg, \ours-YaRN outperforms PoSE-YaRN ~\citep{PoSE} by over 20\% on average in the ``Lost in the Middle''~\citep{Lost_in_the_Middle} task.
    \item \ours can further be enhanced by integrating novel designs on positional interpolation frequencies (such as Linear~\citep{PI}, NTK~\citep{NTK}, Yarn~\citep{YaRN}, \etc), and can be extended to context window sizes of up to 256K or beyond.
    \item \ours-Chat model requires only 100 steps of instruction-tuning to achieve nearly perfect performance on the Needle-in-a-Haystack pressure test, and it outperforms existing strong baselines on LongBench~\citep{longbench}. 

\end{enumerate}

\section{Methodology}
\subsection{Preliminaries}\label{sec:preliminaries}

\paragraph{Problem Formulation.} Given an LLM with a pre-trained context window size $N$, our goal is to unlock the inference capacity of the LLM on the testing data $\mathcal{D}_{\rm test}$ with an extended context window size $L$ (where $L>N$) by \textit{efficiently} learning from a small-scale training data $\mathcal{D}_{\rm train}$ with a maximum sequence length $N$. 
We expect the extended model to perform reasonably well in long-context evaluation. \looseness=-1

\paragraph{Continuity in Positional Encoding.} Transformer-based language models typically encode positional indices sequentially as $\{0,1,\cdots, N-1\}$. Traditional length extension methods~\citep{PI,NTK,YaRN} directly fine-tune on the target length $L$ with an updated positional index. This approach preserves the continuity of all absolute positions and learns all position indices within $[0, L-1]$, thereby successfully extending to the target length. Furthermore, PoSE~\citep{PoSE} attributed their superior performance over RandPos~\citep{randpos} to the ensured continuity of segments during fine-tuning.

\paragraph{Relativity in Positional Encoding.} Relative positional encoding~(RPE)~\citep{RPE} has been proposed as an effective positional encoding method, where only the relative positions between two tokens are considered. Similar to prior works~\citep{randpos,PoSE,skipAlign}, our work focuses on rotary positional encoding~(RoPE)~\citep{su2024roformer}, which is one of the most prominent RPE methods and has been widely applied to LLMs including the recent Llama family~\citep{touvron2023llama,llama2,llama3modelcard}. In RoPE, only the relative distances between position pairs $(|j-i|; 0 \leq i < j \leq L - 1)$ are learned during fine-tuning (\cref{app:rope}). Due to this property, we can manipulate the position indices such that all relative positions between $[0, L-1]$ are learnable within the pre-trained window size. 

\begin{figure}[t!]
    \centering
    \includegraphics[width=1.0\textwidth, height=1.8in]{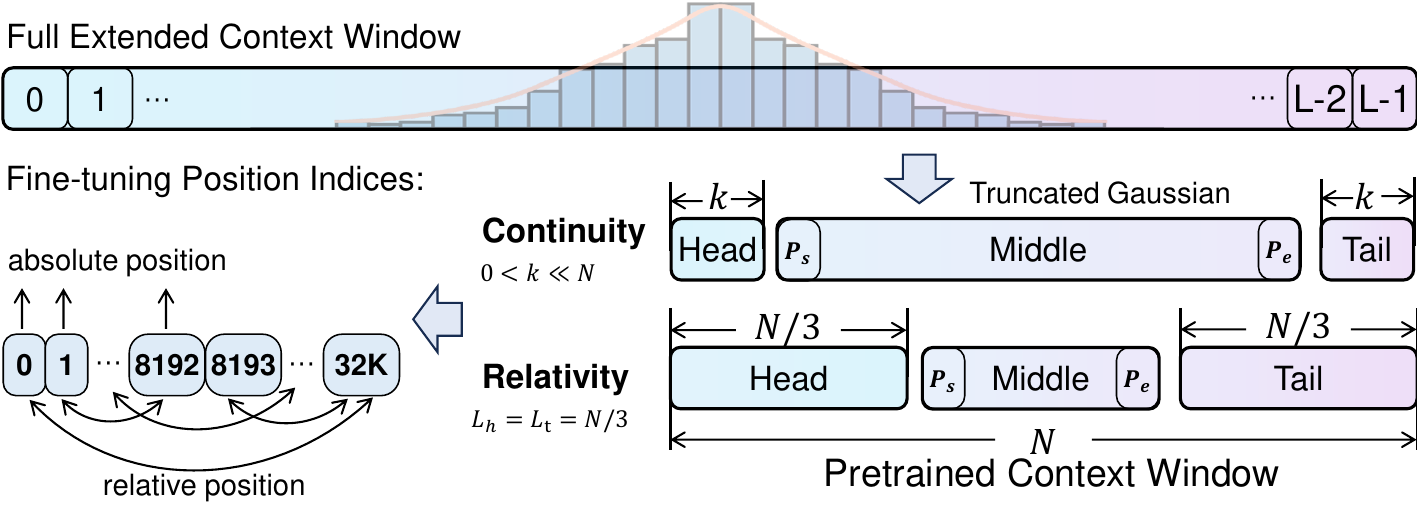}
    \caption{\textbf{Illustration of \ours position interpolation.} The pre-trained context window 
is divided into three segments: the head, middle, and tail. To ensure continuity, we fix the lengths of the head and tail to a small value $k$. To maintain relativity, we set the lengths of the head and tail to $N/3$. For the middle part, the start and end position indices are determined via truncated Gaussian sampling, thereby encouraging the model to pay more attention to the information in the middle part.}
    \label{fig:frame}
\end{figure}

\subsection{Proposed Recipe: \oursfull (\ours)}\label{sec:method}

In the following section, we start by introducing our design of dividing the context window $N$ to learn relative positional information. Then, we propose two strategies that target continuity and relativity, respectively. Lastly, we propose a novel truncated Gaussian sampling method to enhance the middle part of the long context. The overall framework is depicted in \cref{fig:frame}.

\paragraph{Context division.} We first discuss the motivations behind our design of the context length. First, prior works~\citep{lm_infinite,xiao2023efficient} observed that a significant amount of attention score is allocated to the beginning tokens of a sequence, which can potentially encode absolute positional information even without explicit positional encoding~\citep{kazemnejad2024impact}. Secondly, the starting and ending tokens of long contexts can be treated as two pointers that localize the middle indices with the help of relative encodings. Therefore, we divide the pre-trained context window into three segments. The detailed ablation results are shown in \cref{sec:ablation}. \looseness=-1

\begin{definition}\label{def:divsion}
Given the pre-trained context window size $N$ and target extended length $L$, the position set of $\{Head, Middle, Tail\}$ is defined as follows:
\begin{equation}
\label{eq:segments}
\small
    \begin{aligned}
        \text{Head}&=\{0, 1,..., L_h-1\}, \\
        \text{Middle}&=\{P_s,P_s+1,...,P_e-1,P_e\}, \\
        \text{Tail}&=\{L-L_t,...,L-2,L-1\}, \\
        s.t.~~&L_h+(P_e-P_s)+L_t=N,
    \end{aligned}
\end{equation}
where $L_h$ and $L_t$ denote the length of the head and tail segments, $P_s$ and $P_e$ denote the start and end position index of the middle segment.
\end{definition}

The relative positions among the three segments in each sample are calculated in pairs, \ie, $\left\{|j - i|; \forall i,j \in \{Head, Middle, Tail\}\right\}$.

 The formed relative distance union $D_r$ learned by the model is given by:
\begin{equation}
\label{eq:relative_set}
\small
    [0, \max(L_h-1, P_e-P_s, L_t-1)]\cup[P_s-L_h+1, P_e]\cup[L-L_t-P_e, L-1-P_s]\cup[L-L_t-L_h+1, L-1].
\end{equation}
Given that not all samples possess the same values for $L_h$, $P_s$, $P_e$, and $L_t$, as fine-tuning progresses, the union $D_r$ in \cref{eq:relative_set} can encompass the entire range $[0, L-1]$, facilitating the model to learn all relative positions within the target length $L$.

\paragraph{Two segmentation strategies.} For the sake of \textbf{continuity}, we set the $L_h$ and $L_t$ to a very small value $k$, where $0\textless k\ll N$. Specifically, we use $k=32$ in our experiments. This choice allows the middle segment to closely approximate the pre-trained context window. To maintain \textbf{relativity}, we divide $N$ equally into three parts and fix the $L_h$ and $L_t$ to $N/3$, enabling the model to learn as many relative positions as possible. In our fine-tuning process, both types of examples are sampled with equal probability to maintain balance.

\paragraph{Truncated Gaussian Middle Sampling} To better focus the training process on the middle part of the long context, we introduce a truncated Gaussian function. This approach reduces the interval overlap in \cref{eq:relative_set} and directs the model's attention toward the middle section of the long context. In \cref{app:theory}, we provide theoretical justifications of our truncated Gaussian design, indicating that the maximization of $|D_r|$ holds for middle positions in $[N, L/2)\cup (L/2, L-N]$.

Formally, given the probability density function (PDF) of a Gaussian distribution:
\begin{equation}
\nonumber
\small
     f(x) = \frac{1}{\sigma \sqrt{2\pi}} \exp\left(-\frac{(x - \mu)^2}{2\sigma^2}\right),
\end{equation}
where $\mu$ is the mean and $\sigma$ is the standard deviation. The corresponding cumulative distribution function (CDF) is:
\begin{equation}
\label{eq:gau_cdf}
\small
    \begin{aligned}
         F(x) = \int_{-\infty}^{x} f(t) \, dt = 0.5 \left(1 + E \left(\frac{x - \mu}{\sigma \sqrt{2}}\right)\right), \quad{} E(z) = \frac{2}{\sqrt{\pi}} \int_0^z e^{-t^2} \, dt,
    \end{aligned}
\end{equation}
where $E(\cdot)$ is the error function. To calculate the CDF value within the truncated interval, we use a sufficiently large number (\eg 1000) of equally spaced $x$ values from the given interval $[1,L/N]$:
\begin{equation}
\label{eq:gau_x}
\small
     x_i = 1 + \frac{(1 \times (L / N)) \cdot (i - 1)}{999}, \quad i = 1, 2, \ldots, 1000,
\end{equation}
By substituting \cref{eq:gau_x} into \cref{eq:gau_cdf}, the cumulative distribution function (CDF) curve is derived within the truncated interval. For sampling from this truncated Gaussian distribution, the inverse transform method is employed, as demonstrated in \cref{eq:gau_inv}:
\begin{equation}
\label{eq:gau_inv}
\small
     \alpha = \text{round}(x_{i-1} + \frac{(x_i - x_{i-1})(u - F(x_{i-1}))}{F(x_i) - F(x_{i-1})}),
\end{equation}
where $u\sim\text{Uniform}(0,1)$, $\text{round}(\cdot)$ represents rounding to the nearest integer. Finally, we can get:
\begin{equation}
\label{eq:middle}
\small
     \begin{aligned}
         P_e &\sim \text{Uniform}(L_h + \alpha \times L_m, (\alpha \times N - 1) - L_t), \\
         P_s &= P_e - L_m + 1,
     \end{aligned}
\end{equation}
where $L_m$ denotes the length of the middle segments. In summary, the overall sampling flow of our algorithm is presented in \cref{alg:algorithm}.
\begin{algorithm}[!ht]
\small
\caption{\ours sampling algorithm}
\label{alg:algorithm}
\begin{algorithmic}[1]
\REQUIRE Pre-trained context window size $N$, extended context window size $L$, training sample size $S$, mean $\mu$, variance $\sigma$ and hyperparameter $k$.
\STATE Generate enough $x$ equally spaced according to Equation~\eqref{eq:gau_x}.
\STATE Substitute $x$ into Equation~\eqref{eq:gau_cdf} to derive the truncated Gaussian CDF $F(x)$.
\FOR{$i = 0$ to $S-1$}
    \STATE Sample $L_h \sim \text{DiscreteUniform}(\{k, N/3\})$, and let $L_t=L_h, L_m=N-L_h-L_t$.
    \STATE Sample $u\sim\text{Uniform}(0,1)$, and substitute it into Equation~\eqref{eq:gau_inv} to get $\alpha$.
    \STATE Calculate the start and end position ids $P_s,P_e$ of the middle part according to Equation~\eqref{eq:middle}.
    \STATE Get position set $P_i=\{0, 1, \ldots, L_h, P_s, \ldots, P_e, L-L_t \ldots, L-1\}$, where $|P|=N$. 
\ENDFOR
\RETURN $P=\{P_0, P_1, \ldots, P_{S-2}, P_{S-1}\}$.
\end{algorithmic}
\end{algorithm}

\section{Experiments}\label{sec:experiments}

\subsection{Experimental Setup}

\paragraph{Extended Models}
We use \texttt{Llama-2-7B} and \texttt{Llama-2-7B-Chat}~\citep{llama2} as the base models and extend their pre-trained context window size of 4K to a target context length of 32K. The extended models are referred to as \ours-Base and \ours-Chat, respectively. Note that, though the target context length is 32K, we do not have to fine-tune \ours on 32K token long text (see Section~\ref{sec:method}). 

\paragraph{Benchmarks}
We conduct long-context LLM evaluation of \ours-Base on LongChat-Lines~\citep{Giraffe} and Lost-in-the-Middle~\citep{Lost_in_the_Middle}. Ideally, fine-tuning should not disrupt what the base model has learned, so we further evaluate \ours-Base on the language modeling task and the evaluation benchmark~\citep{open_llm_leaderboard} adopted by Llama2. Additionally, we assess the \ours-Chat model with Needle-in-a-Haystack\footnote{\url{https://github.com/gkamradt/LLMTest_NeedleInAHaystack}} and LongBench\citep{longbench}. Unless otherwise specified, we use linear interpolation to adapt LLMs to a longer context length.

% 已更新：加baselines
\paragraph{Baselines}
As far as we know, RandPos~\citep{randpos} and PoSE~\citep{PoSE} are similar to our approach in that they manipulate position indices to enable fine-tuning on the pre-trained length for context expansion. Therefore, these two methods serve as the baselines for our primary comparisons. More details about the experimental setup can be found in the \cref{app:exp_details}.

\subsection{Effective Context Window Size Evaluation on \ours-Base}

\begin{figure}[htbp]
    \centering
    \includegraphics[width=1.0\textwidth]{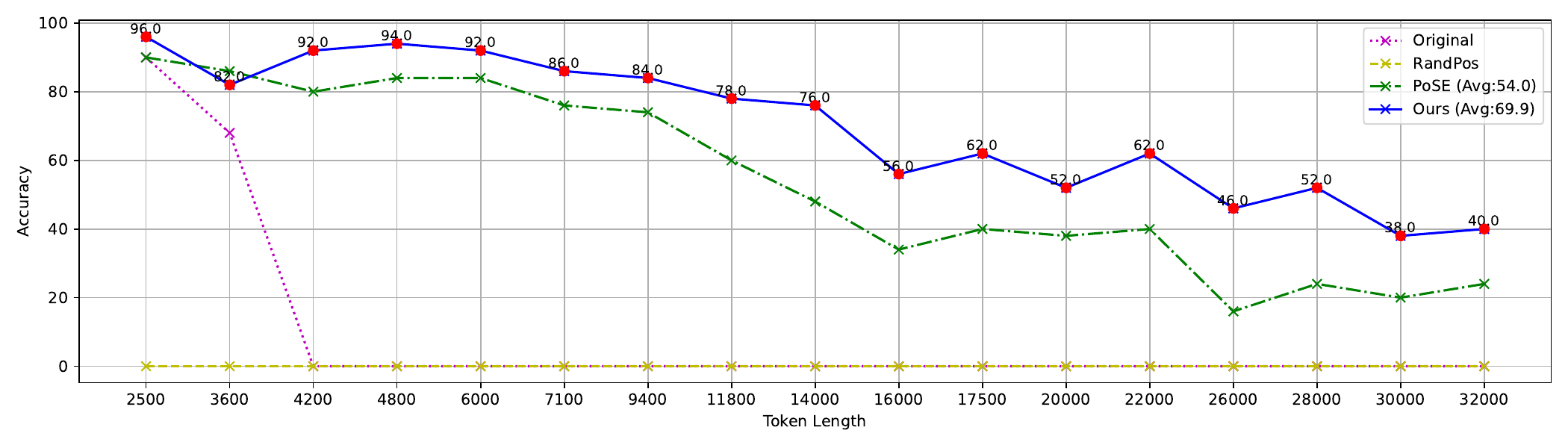}
    \caption{\textbf{Results (\%) on LongChat-Lines}. Each length consists of 50 samples. All results are fine-tuned on \texttt{Llama-2-7B} with 4K length data through linear position interpolation. Refer to \cref{app:longchat_lines} for ablated results using NTK~\citep{NTK} and Yarn~\citep{YaRN}.}
    \label{fig:longchat_lines}
\end{figure}

We evaluate the long-context understanding capabilities of the \ours-Base model on two tasks: LongChat-Lines\footnote{Passkey retrieval~\citep{passkey} is another similar task for evaluating long-context LLMs, but it is too simplistic to reflect model performance at different context window sizes, so we use the dataset provided by \citet{Giraffe}, which closely aligns with the task described in \citet{Longchat}}~\citep{Giraffe} (\cref{fig:longchat_lines}) and ``Lost in the Middle''~\citep{Lost_in_the_Middle} (\cref{tab:lost_in_the_middle}). 

% \begin{figure}[htbp]
%     \centering
%     \begin{subfigure}[b]{1.0\textwidth}
%         \includegraphics[width=\textwidth]{figure/crop_Linear.pdf}
%         \caption{First subfigure}
%         \label{fig:sub1}
%     \end{subfigure}

%     \begin{subfigure}[b]{1.0\textwidth}
%         \includegraphics[width=\textwidth]{figure/crop_NTK.pdf}
%         \caption{Second subfigure}
%         \label{fig:sub2}
%     \end{subfigure}

%     \begin{subfigure}[b]{1.0\textwidth}
%         \includegraphics[width=\textwidth]{figure/crop_Yarn.pdf}
%         \caption{Third subfigure}
%         \label{fig:sub3}
%     \end{subfigure}

%     \caption{Three vertically aligned subfigures}
%     \label{fig:test}
% \end{figure}

\begin{table}[h!]
    \centering
    % \small
    % \setlength\tabcolsep{5pt} % 列间距
    \caption{\textbf{Results (\%) on ``Lost in the Middle''}. ``Position'' indicates the correct answers' index, and each index comprises 500 samples. All results are fine-tuned on \texttt{Llama-2-7B} with 4K length data.}
    \resizebox{\linewidth}{!}{
    \begin{tabular}{lcccccc|ccccccc}
    \toprule
    \multirow{3}{*}{\textbf{Model}} & \multicolumn{5}{c}{\textbf{Position} (75 keys, $\sim$\textbf{5K} tokens)} & \multirow{3}{*}{\textbf{AVG}} & \multicolumn{5}{c}{\textbf{Position} (140 keys, $\sim$\textbf{10K} tokens)} & \multirow{3}{*}{\textbf{AVG}} \\
    \cmidrule(lr){2-6} \cmidrule(lr){8-12}
    & \textbf{0} & \textbf{18} & \textbf{37} & \textbf{54} & \textbf{74} & & \textbf{0} & \textbf{34} & \textbf{69} & \textbf{104} & \textbf{139} & \\
    \midrule
    PoSE-Linear & 99.4 & 24.4 & 37.4 & 47.2 & 46.2 & 50.9 & 95.2 & 8.2 & 7.6 & 13.8 & 18.6 & 28.7 \\
    \ours{}-Linear & 99.6 & \cellcolor[HTML]{96FFFB}{45.6} & \cellcolor[HTML]{96FFFB}{56.0} & \cellcolor[HTML]{96FFFB}{67.0} & 58.0 & \textbf{65.2} & 96.6 & \cellcolor[HTML]{96FFFB}{19.8} & \cellcolor[HTML]{96FFFB}{23.4} & \cellcolor[HTML]{96FFFB}{31.0} & 26.2 & \textbf{39.4} \\ \midrule
    PoSE-NTK & 98.6 & 49.6 & 44.6 & 40.2 & 41.4 & 54.9 & 97.6 & 3.4 & 0 & 0 & 27.6 & 25.7 \\
    \ours{}-NTK & 96.2 & \cellcolor[HTML]{96FFFB}{53.8} & \cellcolor[HTML]{96FFFB}{52.6} & \cellcolor[HTML]{96FFFB}{72.8} & 42.0 & \textbf{63.5} & 78.6 & \cellcolor[HTML]{96FFFB}{5.2} & \cellcolor[HTML]{96FFFB}{6.0} & \cellcolor[HTML]{96FFFB}{23.4} & 41.8 & \textbf{29.9} \\ \midrule
    PoSE-YaRN & 99.6 & 32.6 & 12.2 & 57.2 & 48.4 & 50.0 & 91.8 & 0.6 & 2.8 & 8.2 & 18.8 & 24.4 \\
    \ours{}-YaRN & 100.0 & \cellcolor[HTML]{96FFFB}{49.6} & \cellcolor[HTML]{96FFFB}{47.6} & \cellcolor[HTML]{96FFFB}{77.4} & 92.6 & \textbf{73.4} & 99.4 & \cellcolor[HTML]{96FFFB}{8.0} & \cellcolor[HTML]{96FFFB}{5.8} & \cellcolor[HTML]{96FFFB}{43.8} & 69.2 & \textbf{45.2} \\
    \bottomrule
     \label{tab:lost_in_the_middle}
    \end{tabular}
    }
\end{table}

\paragraph{\ours-Base performs best in retrieving information from long contexts of varying lengths.} We extend the context window size up to 32K and compare \ours with the \texttt{Llama 2-7B}~\citep{llama2}, RandPos~\citep{randpos}, and PoSE~\citep{PoSE}. As the context window size increases, the performance of all models drops, but \ours always performs best except for the window size of 3.6K (see \cref{fig:longchat_lines}). In terms of the average performance over all context window sizes, \ours outperforms PoSE by 16\%, demonstrating its good long-context understanding ability.

\paragraph{\ours-Base alleviates the Lost-in-the-Middle issue.} Lost-in-the-Middle is an observation that LLMs are generally good at retrieving relevant information appearing at the beginning/end of the input context~\citep{Lost_in_the_Middle}. To validate the effectiveness of our middle-focused truncated Gaussian sampling, we evaluate \ours and compare it with PoSE on the key-value retrieval task proposed by~\citet{Lost_in_the_Middle}. 
We present results in \cref{tab:lost_in_the_middle}, where the cyan shading indicates middle segments. We find that: regardless of the chosen interpolation method, \ours always outperforms PoSE by a large margin. \eg, \ours-Linear surpasses PoSE-Linear by 21.2\% when the relevant information is placed at 18.

\subsection{Long Context Understanding Evaluation on \ours-Chat}

We conduct long-context evaluations of \ours-Chat on two tasks: 

\begin{itemize}[leftmargin=*,noitemsep,topsep=0pt]
    \item\textbf{Needle in A Haystack} (\cref{fig:needle})\quad{} This task is a test that places an answer (\ie, Needle) at any position of a long context window (\ie, Haystack) and requires a model to retrieve the correct answer given a question-answer pair. We follow \citet{skipAlign} and use the GPT (\texttt{GPT-3.5-Turbo-0125}) score as the evaluation metric.
    \item \textbf{LongBench} (\cref{tab:longbench2})\quad{} \citet{longbench} is a more realistic benchmark because it covers real-world application scenarios like long-context QA and summarization. Moreover, it is specifically designed for Chat models.
\end{itemize}

% \begin{figure}[htbp]
%     \centering
%     \includegraphics[width=0.8\textwidth]{figure/crop_needle.pdf}
%     \caption{Needle}
%     \label{fig:needle}
% \end{figure}

\begin{figure}[h]
    \centering
    \begin{subfigure}{0.48\textwidth}
        \centering
        \includegraphics[width=\linewidth, height=1.8in]{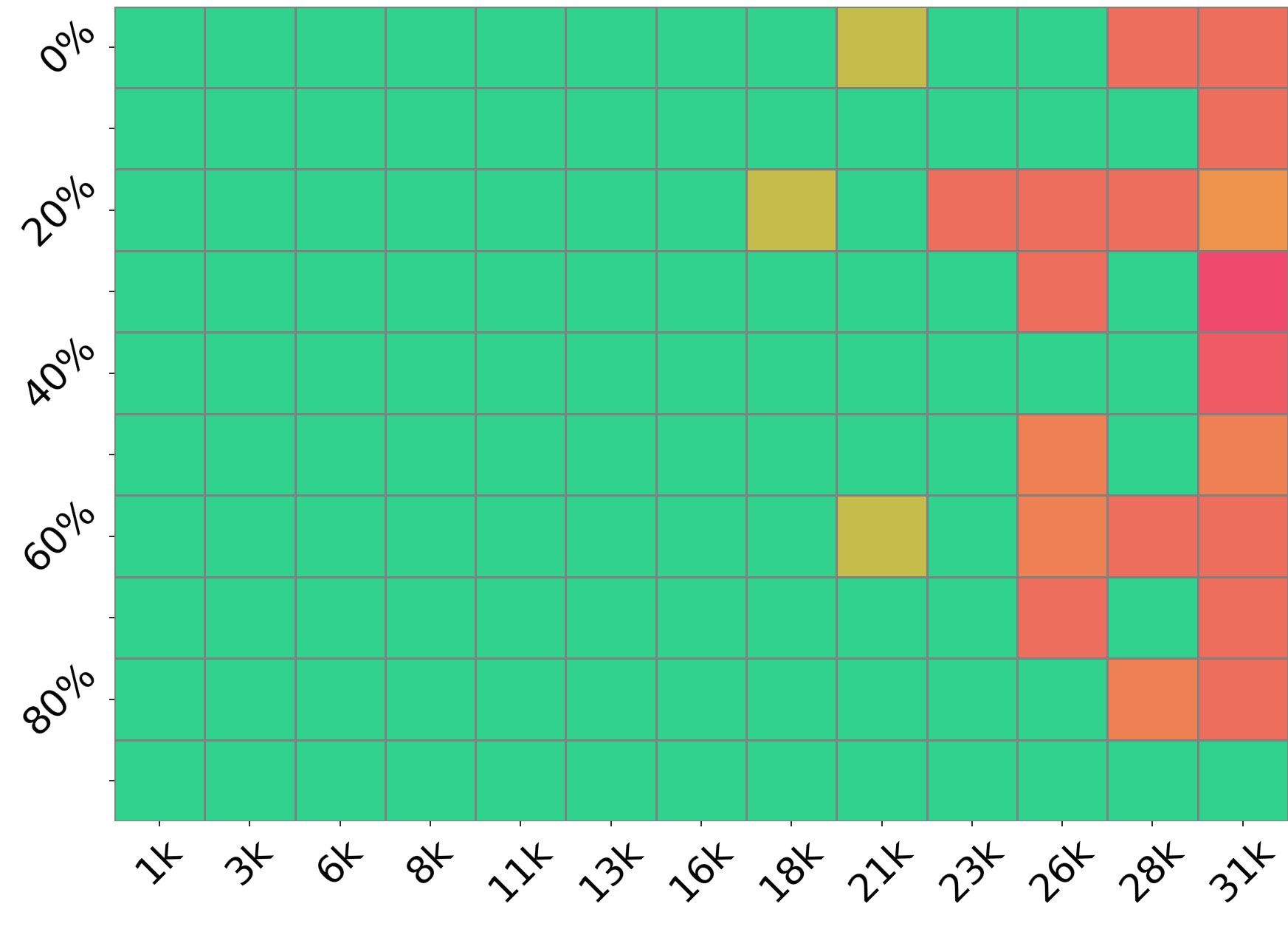}
        \caption{SkipAlign$^\dag$}
    \end{subfigure}
    \hfill
    \begin{subfigure}{0.48\textwidth}
        \centering
        \includegraphics[width=\linewidth, height=1.8in]{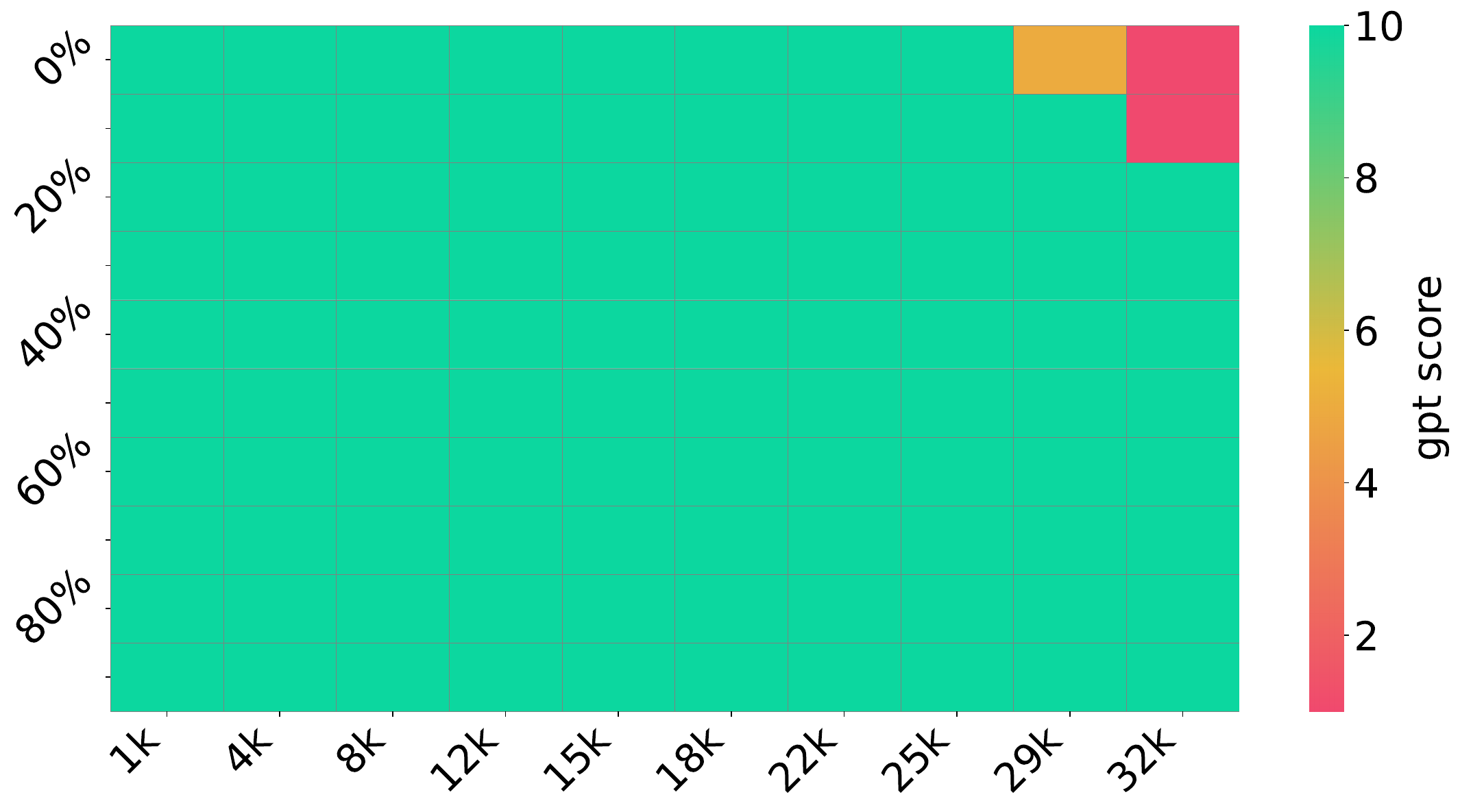}
        \caption{\ours{}}
    \end{subfigure}
    \caption{\textbf{Results on Needle-in-a-Haystack.} $^\dag$ indicates the results excerpted from \citet{skipAlign}. Both results are instruction-tuned on \texttt{LLaMa2-7B-Chat} with 4K length data. The color gradually changes from deep green to deep red, indicating the Recall performance decreases from 10 to 1. }
    \label{fig:needle}
\end{figure}

\begin{table}[!ht]
\centering
% \small
% \setlength\tabcolsep{4pt} % 列间距
\caption{\textbf{Results (\%) on LongBench}. $^*$ indicates results reported by \citet{longbench}. \ours{}-7B-32k is instruction-tuned for 100 steps using 4K length data on \texttt{LLaMa2-7B-Chat}.}
\resizebox{\linewidth}{!}{
\begin{tabular}{lccccccc}
\toprule
\textbf{Model} & \textbf{\begin{tabular}[c]{@{}l@{}}Single-\\ Doc QA\end{tabular}}  & \textbf{\begin{tabular}[c]{@{}c@{}}Multi-\\ Doc QA\end{tabular}} & \textbf{\begin{tabular}[c]{@{}c@{}}Summari-\\ zation\end{tabular}} & \textbf{\begin{tabular}[c]{@{}c@{}}Few-shot\\ Learning\end{tabular}} & \textbf{\begin{tabular}[c]{@{}c@{}}Code\\ Completion\end{tabular}} & \textbf{\begin{tabular}[c]{@{}c@{}}Synthetic\\ Tasks\end{tabular}} & \textbf{AVG} \\
\midrule
Llama2-7B-chat-4k$^*$ & 24.9 & 22.6 & 24.7 & 60.0 & 48.1 & 5.9 & 31.0 \\
XGen-7B-8k$^*$ & 24.6 & 20.4 & 24.7 & 56.2 & 38.6 & 5.3 & 28.3 \\
Mistral-7B-Instruct-v0.1 & 29.5 & 20.7 & 26.4 & 13.6 & 29.6 & 10.8 & 21.8 \\
Mistral-7B-Instruct-v0.2 & 28.5 & 21.5 & 26.1 & 50.1 & 33.8 & 13.9 & 29.0 \\
Mistral-7B-Instruct-v0.3 & 33.2 & 30.6 & 26.8 & 56.4 & 15.3 & 10.4 & 28.8 \\
InternLM-7B-8k$^*$ & 17.4 & 20.2 & 16.1 & 50.3 & 36.4 & 4.5 & 24.2 \\
Vicuna-v1.5-7B-16k$^*$ & 28.0 & 18.6 & 26.0 & 66.2 & 47.3 & 5.5 & 31.9 \\
LongChat-v1.5-7B-32k$^*$ & 28.7 & 20.6 & 26.7 & 60.0 & 54.1 & 15.8 & 34.3 \\ \midrule
\ours-7B-32k & 34.8 & 31.1 & 27.2 & 65.1 & 50.4 & 7.0 & \cellcolor[HTML]{FFFC9E}{\textbf{35.9}} \\
\bottomrule
\label{tab:longbench2}
\end{tabular}
}
\end{table}

% \footnotetext{The detailed sub-task results can be found in Appendix~\ref{app:longbench_subtasks}.}

\paragraph{\ours-Chat outperforms SkipAlign in context window expansion.} 
We visualize the results of \ours-Chat and the recent SkipAlign in Figure~\ref{fig:needle}. Clearly, \ours-Chat beats SkipAlign because the performance of SkipAlign~\citep{skipAlign} decreases from the window size of 18K while \ours-Chat displays a perfect performance everywhere until from the window size of 29K. Notably, \ours-Chat is only fine-tuned for 100 steps.

\paragraph{\ours-Chat makes best use of the extended context window size.} 
We present results on LongBench in \cref{tab:longbench2}. \ours-Chat again surpasses strong baseline models, demonstrating its better use of extended context size. In terms of the average performance over all tasks, it outperforms the second best model, \ie, \texttt{LongChat-v1.5-7B-32k}~\citep{Longchat}, by 1.6\%, though it is only tuned on a very small amount of data and for only 100 steps.

\subsection{Effectiveness of PEFT Integration}

To demonstrate that \ours can be directly combined with PEFT techniques (such as LoRA~\citep{lora} and QLoRA~\citep{qlora}), requiring no additional modifications. We conducted experiments on \texttt{LLaMa-2-7B-Chat} using the identical dataset and settings. The experimental results are presented in Table~\ref{tab:lora}. The results indicate that models fine-tuned using LoRA and QLoRA achieve performance nearly equivalent to those fine-tuned with full parameter. 
% This compatibility stems from the fact that \ours does not modify data formats or model architectures during the fine-tuning process; instead, it operates solely by adjusting position indices.

\begin{table}[htbp]
    \centering
    \small
    \caption{\textbf{Results (\%) on LongBench}. $^*$ indicates results reported by \citet{longbench}. \ours{}-7B-32k is instruction-tuned for 100 steps using 4K length data on \texttt{LLaMa2-7B-Chat}.}
    \resizebox{\linewidth}{!}{
    \begin{tabular}{@{}lccccccc@{}}
    \toprule
        \multirow{2}{*}{\textbf{Model}} & \multirow{2}{*}{\textbf{\begin{tabular}[c]{@{}c@{}}Single-\\ Doc QA\end{tabular}}} & \multirow{2}{*}{\textbf{\begin{tabular}[c]{@{}c@{}}Multi-\\ Doc QA\end{tabular}}} & \multirow{2}{*}{\textbf{\begin{tabular}[c]{@{}c@{}}Summari-\\ zation\end{tabular}}} & \multirow{2}{*}{\textbf{\begin{tabular}[c]{@{}c@{}}Few-shot\\ Learning\end{tabular}}} & \multirow{2}{*}{\textbf{\begin{tabular}[c]{@{}c@{}}Code\\ Completion\end{tabular}}} & \multirow{2}{*}{\textbf{\begin{tabular}[c]{@{}c@{}}Synthetic\\ Tasks\end{tabular}}} & \multirow{2}{*}{\textbf{Macro}} \\
         &  &  &  &  &  &  &  \\ \midrule
        Llama2-7B* & 24.9 & 22.6 & 24.7 & 60.0 & 48.1 & 5.9 & 31.0 \\
        LoRA & 28.9 & 28.6 & 27.8 & 62.2 & 54.6 & 10.8 & 35.5 \\
        QLoRA & 28.1 & 27.6 & 28.1 & 61.7 & 54.6 & 10.1 & 35.0 \\
        \ours-7B-32k     & 34.8 & 31.1 & 27.2 & 65.1 & 50.4 & 7.0 & 35.9 \\
    \bottomrule
    \label{tab:lora}
    \end{tabular}
    }
\end{table}

\subsection{Language Modeling and Standard Benchmark}

Following~\citet{PI, PoSE, YaRN}, we perform the classic language modeling evaluation, \ie, perplexity evaluation, on GovReport~\citep{gov_R} and Proof-pile~\citep{proof_pile}. Since a lower perplexity does not necessarily imply better model performance on downstream tasks~\citep{EABF,ppl_lm,arora2024zoology,park2024can}, we further conduct evaluation on the standard natural-language-understanding (NLU) benchmark~\citep{open_llm_leaderboard}. This also lets us know whether fine-tuning hurts the NLU ability of the pre-trained base model.

 \begin{table}[htbp]
    \vspace{-2.0em}
    \centering
    \small
    \setlength\tabcolsep{12.9pt} % 列间距
    \caption{\textbf{Perplexity results of GovReport and Proof-pile.} Each experiment is the average perplexity of 50 samples, and all results are based on \texttt{LLaMa2-7B} fine-tuned on 4K data length.}
    \resizebox{\linewidth}{!}{
    \begin{tabular}{@{}lcccccccccc@{}}
    \toprule
    % & \multirow{3}{*}{\begin{tabular}[c]{@{}c@{}}\textbf{Context Size}\\ \textbf{Train / Target}\end{tabular}}
    \multirow{3}{*}{\textbf{Model}} & \multicolumn{4}{c}{\textbf{GovReport}} & \multicolumn{4}{c}{\textbf{Proof-pile}} \\
    \cmidrule(lr){2-5} \cmidrule(lr){6-9}
    & \textbf{4K} & \textbf{8K} & \textbf{16K} & \textbf{32K} & \textbf{4K} & \textbf{8K} & \textbf{16K} & \textbf{32K} \\ \midrule
    Original & 3.6 & - & - & - & 4.6 & - & - & - \\ \midrule
    RandPos-Linear &  8.9 & 7.4 & 6.2 & 5.8 & 12.1 & 11.9 & 11.9 & 12.9 \\
    PoSE-Linear &  3.8 & 3.2 & 2.7 & 2.5 & 4.7 & 4.6 & 4.6 & 4.4 \\
    \ours{}-Linear &  3.8 & 3.2 & 2.7 & 2.5 & 4.7 & 4.6 & 4.5 & 4.4 \\ \midrule
    RandPos-NTK &  4.6 & 4.0 & 3.6 & 4.0 & 5.8 & 5.8 & 6.2 & 7.3 \\
    PoSE-NTK &  3.7 & 3.2 & 2.7 & 2.6 & 4.7 & 4.6 & 4.5 & 4.7 \\
    \ours{}-NTK &  3.8 & 3.2 & 2.7 & 2.7 & 4.7 & 4.6 & 4.5 & 4.7 \\ \midrule
    RandPos-YaRN &  5.0 & 4.4 & 4.0 & 4.6 & 6.4 & 6.5 & 6.8 & 9.1 \\
    PoSE-YaRN &  3.7 & 3.2 & 2.7 & 2.5 & 4.6 & 4.6 & 4.5 & 4.4 \\
    \ours{}-YaRN &  3.7 & 3.2 & 2.7 & 2.5 & 4.6 & 4.6 & 4.5 & 4.4 \\
    \bottomrule
    \label{tab:linear_ppl}
    \end{tabular}
    }
    \vspace{-2.0em}
\end{table}

\paragraph{Both \ours and PoSE demonstrate the lowest perplexity.} We apply different positional interpolation methods to RandPos~\citep{randpos}, PoSE~\citep{PoSE}, and \ours and report their perplexities in~\cref{tab:linear_ppl}. We find that: \ours and PoSE have a similar perplexity in different settings and both outperform RandPos. This occurs primarily because the position indices used during RandPos fine-tuning are discontinuous, which creates an inconsistency with the pre-training stage.

\begin{table}[htbp]
    \centering
    % \tiny
    % \setlength\tabcolsep{8.2pt} % 列间距
    \caption{\textbf{Experimental results of standard benchmarks.} $^*$ indicates results cited from \citet{llama2}, and all results are based on \texttt{LLaMa2-7B} fine-tuned on 4K data length.}
    \resizebox{\linewidth}{!}{
    \begin{tabular}{@{}lcccccc@{}}
    \toprule
    \multirow{3}{*}{\textbf{Model}} & \multicolumn{4}{c}{\textbf{Zero-Shot}} & \multicolumn{2}{c}{\textbf{Few-Shot}} \\
    \cmidrule(lr){2-5} \cmidrule(lr){6-7}
    & \textbf{WinoGrande} & \textbf{TruthfulQA(mc2)} & \textbf{PIQA} & \textbf{BoolQ} & \textbf{ARC-C} & \textbf{HellaSwag} \\
    \midrule
    LLaMa-2-7b-hf$^*$& 69.2 & 39.5 & 78.8 & 77.4 & 45.9 & 77.2 \\ \midrule
    RandPos-Linear & 63.3 & 39.3 & 76.5 & 66.6 & 32.0 & 48.5 \\
    PoSE-Linear & 68.8 & 38.6 & 77.8 & 76.2 & 47.7 & 77.1 \\
    \ours{}-Linear & 67.5 & 37.4 & 78.5 & 75.4 & 46.8 & 76.9 \\ \midrule
    RandPos-NTK & 68.7 & 35.9 & 78.6 & 74.8 & 45.5 & 74.4 \\
    PoSE-NTK & 68.8 & 38.6 & 77.8 & 76.2 & 47.7 & 77.1 \\
    \ours{}-NTK & 67.5 & 37.4 & 78.5 & 75.4 & 46.8 & 76.9 \\ \midrule
    RandPos-YaRN & 69.3 & 36.6 & 78.3 & 72.5 & 43.4 & 69.2 \\
    PoSE-YaRN & 69.4 & 39.6 & 78.1 & 76.7 & 49.0 & 78.0 \\
    \ours{}-YaRN & 68.7 & 38.5 & 78.0 & 76.4 & 49.0 & 78.0 \\
    \bottomrule
    \label{tab:standard_benchmark}
    \end{tabular}
    }
    \vspace{-2.0em}
\end{table}

\paragraph{\ours has nearly the same NLU abilities as the pre-trained base model.} Ideally, fine-tuning should not adversely affect the original capabilities of the pre-trained base model. Our evaluation of \ours confirms this, \ie, \ours nearly retains all NLU abilities of the base Llama2-7B (see \cref{tab:standard_benchmark}). Interestingly, \ours improves over Llama2-7B on ARC-C and HellaSwag. This is because these two tasks are few-shot tasks with longer prompts, necessitating the assistance of long-context understanding. 
% Again, PoSE~\citep{PoSE} performs similarly to \ours and both beat RandPos~\citep{randpos}.

\paragraph{Extending the context length to 256K.}

We push the limit and extend the context length of \texttt{Llama-2-7B} up to 256K. Following~\citet{PoSE}, we evaluate the extended model by calculating the average perplexity over 20 samples from PG-19~\citep{pg19} and Book3~\citep{book3}.\footnote{We use sliding window for calculation, with a window size of 32,768 and a sliding step size of 4,096.} Since the PG-19 test set does have enough samples that are longer than 256K, we select a subset of samples from the PG-19 training set.

\begin{table}[htbp]
    \centering
    % \small
    % \setlength\tabcolsep{7pt} % 列间距
    \caption{\textbf{Perplexity results of PG-19 and Book3.} $^*$ indicates results copied from \citet{PoSE}, and \ours{} is based on \texttt{LLaMa2-7B} fine-tuned on 4K data length. }
    \resizebox{\linewidth}{!}{
    \begin{tabular}{@{}lcccccccccc@{}}
    \toprule
    \multirow{3}{*}{\textbf{Model}} & \multicolumn{5}{c}{\textbf{PG-19}} & \multicolumn{5}{c}{\textbf{Book3}} \\
    \cmidrule(lr){2-6} \cmidrule(lr){7-11}
    & \textbf{64K} & \textbf{96K} & \textbf{128K} & \textbf{192K} & \textbf{256K} & \textbf{64K} & \textbf{96K} & \textbf{128K} & \textbf{192K} & \textbf{256K} \\
    \midrule
    PoSE-Linear-128K$^*$ & 22.47 & 26.77 & 31.18 & - & - & 43.62 & 57.08 & 70.87 & - & - \\
    PoSE-NTK-128K$^*$ & 14.84 & 29.48 & 34.80 & - & - & 16.04 & 31.42 & 37.00 & - & - \\
    PoSE-YaRN-128K$^*$ & 10.36 & 10.77 & 11.33 & - & - & 12.30 & 13.07 & 13.81 & - & - \\ \midrule
    \ours{}-Linear-192K & 5.9 & 6.0 & 6.1 & 6.1 & - & 7.6 & 7.7 & 7.8 & 7.8 & -  \\
    \ours{}-NTK-192K & 5.0 & 5.1 & 5.2 & 5.2 & - & 6.9 & 7.0 & 7.0 & 7.1 & - \\
    \ours{}-YaRN-192K & 5.0 & 5.2 & 5.2 & 5.3 & - & 7.0 & 7.1 & 7.1 & 7.1 & -  \\ \midrule
    \ours{}-Linear-256K & 7.8 & 8.0 & 8.0 & 8.1 & 8.2 & 10.2 & 10.3 & 10.5 & 10.7 & 10.8  \\
    \ours{}-NTK-256K & 5.1 & 5.3 & 5.3 & 5.4 & 5.4 & 7.2 & 7.3 & 7.3 & 7.3 & 7.4   \\
    \ours{}-YaRN-256K & 5.2 & 5.3 & 5.4 & 5.4 & 5.5 & 7.1 & 7.2 & 7.2 & 7.3 & 7.3   \\
    \bottomrule
    \label{tab:extremely_long}
    \end{tabular}
    }
\end{table}

We experiment with target context lengths 64K, 96K, 128K, 192K, and 256K and apply different positional interpolation methods to the extended model (see \cref{tab:extremely_long}). The results of PoSE~\citep{PoSE} in \cref{tab:extremely_long} are based on fine-tuning \texttt{LLaMa 1-7B} with 2K data length, and are provided for reference only. Surprisingly, the increase of the target context length brings little to no perplexity increase, demonstrating the stability of \ours across different target context lengths, even when the target context is extremely long.

\subsection{Ablation Study}\label{sec:ablation}

To validate the effectiveness of our modeling choices, we further conduct an ablation study of three main components of \ours: 
% Furthermore, to demonstrate the effectiveness of each component of \ours, we conduct experiments focusing on three aspects: 
truncated Gaussian sampling, fixed start and end segments, and the trade-off between continuity and relativity.

\begin{figure}[h]
    \centering
    \begin{subfigure}{0.32\textwidth}
        \centering
        \includegraphics[width=\linewidth]{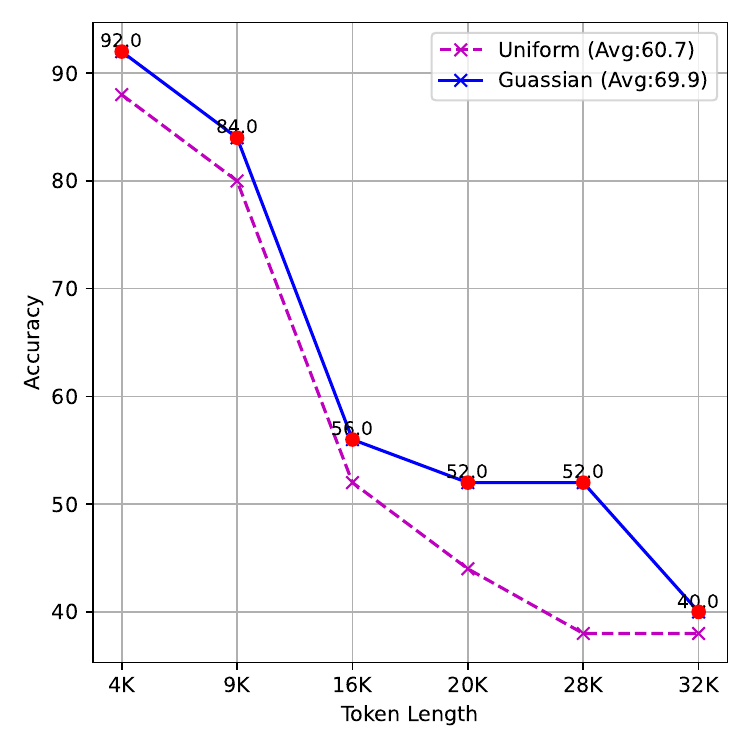}
        \caption{Gaussian \vs Uniform.}
    \end{subfigure}
    \hfill
    \begin{subfigure}{0.32\textwidth}
        \centering
        \includegraphics[width=\linewidth]{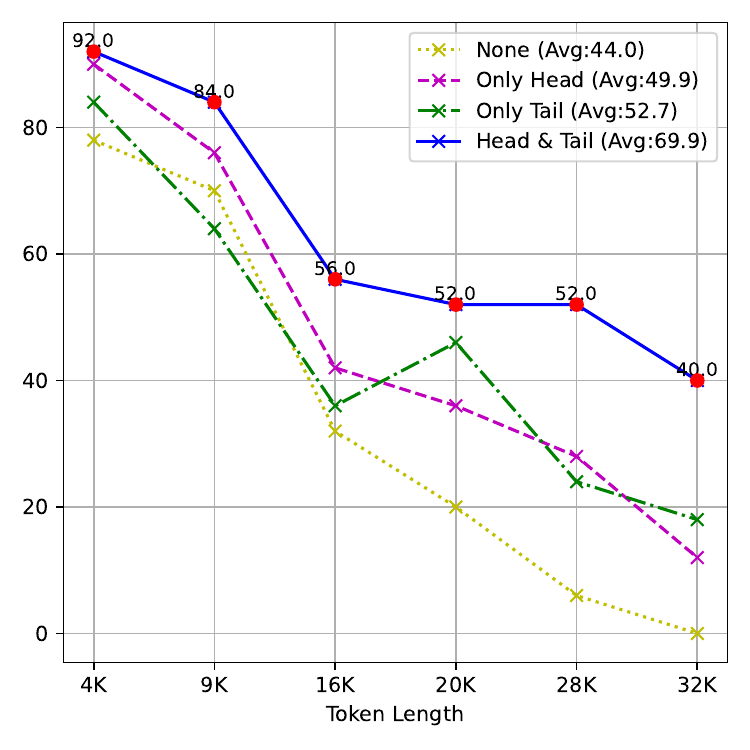}
        \caption{Head \vs Tail.}
    \end{subfigure}
    \hfill
    \begin{subfigure}{0.32\textwidth}
        \centering
        \includegraphics[width=\linewidth]{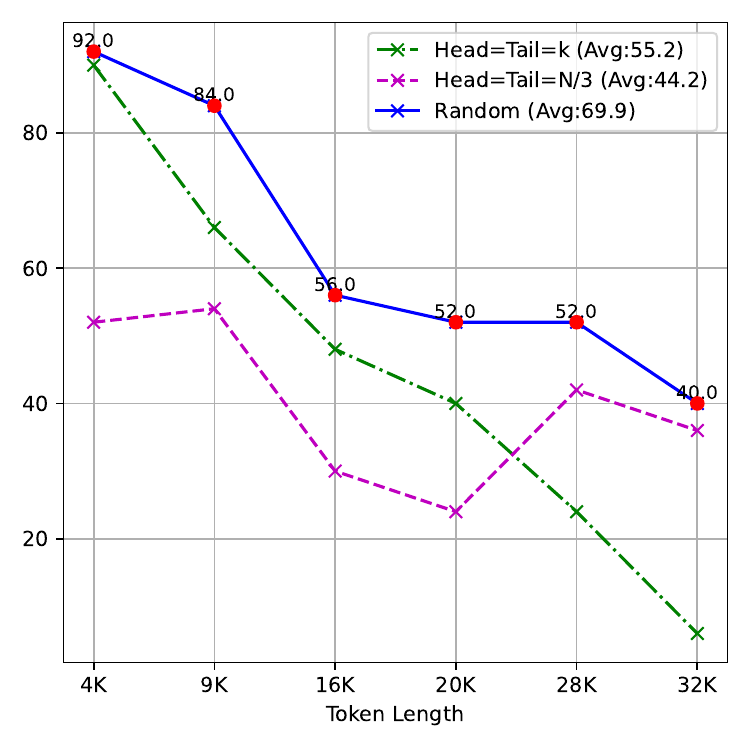}
        \caption{Relativity \vs Continuity.}
    \end{subfigure}
    \caption{\textbf{Ablation study of \ours on LongChat-Lines}. The result at each length is estimated using 50 samples.
    % Each length consisting of 50 samples.
    }
    \label{fig:ablation}
\end{figure}

\paragraph{Truncated Gaussian sampling versus Uniform sampling.}
We use truncated Gaussian sampling to encourage \ours to make better use of the middle part of the context. As a comparison, we replace it with the Uniform sampling (see Figure~\ref{fig:ablation}(a)). We observe that the Uniform sampling always leads to worse retrieval performance, suggesting the effectiveness of the truncated Gaussian sampling.

\paragraph{Fixing the head and tail segments is crucial for good retrieval performance.} 
We compare our choice of fixing the head and tail segments with three alternatives: (i) removing both the head and tail segment, (ii) fixing only the head segment, and (iii) fixing only the tail segment (see Figure~\ref{fig:ablation}(b)). 
We find that: removing the head and tail segments leads to the worst performance; it results in a complete failure (\ie, zero score) for the context size 32K. Keeping either head or tail segments performs slightly better than removing both but underperforms our default choice of fixing both. We suppose that this is because fixing both gives rise to better relativity information, a finding that is consistent with that of~\citet{lm_infinite}.

\paragraph{Maintaining a good balance between continuity and relativity is necessary.}
We encourage continuity by setting the head and tail segment lengths to $k=32$ and elicit relativity by letting $k=N/3$ (see Section~\ref{sec:method}). To balance the two desired properties, we randomly choose $k=32$ and $k=N/3$ with an equal probability during fine-tuning. Here we compare three scenarios: (1) enforce only continuity, (2) enforce only relativity, and (3) balance continuity and relativity (see Figure~\ref{fig:ablation}(c)). We find that balancing continuity and relativity gives rise to the best performance, thus justifying our modeling choice.

\begin{figure}[htbp]
    \centering
    \includegraphics[width=1.0\textwidth]{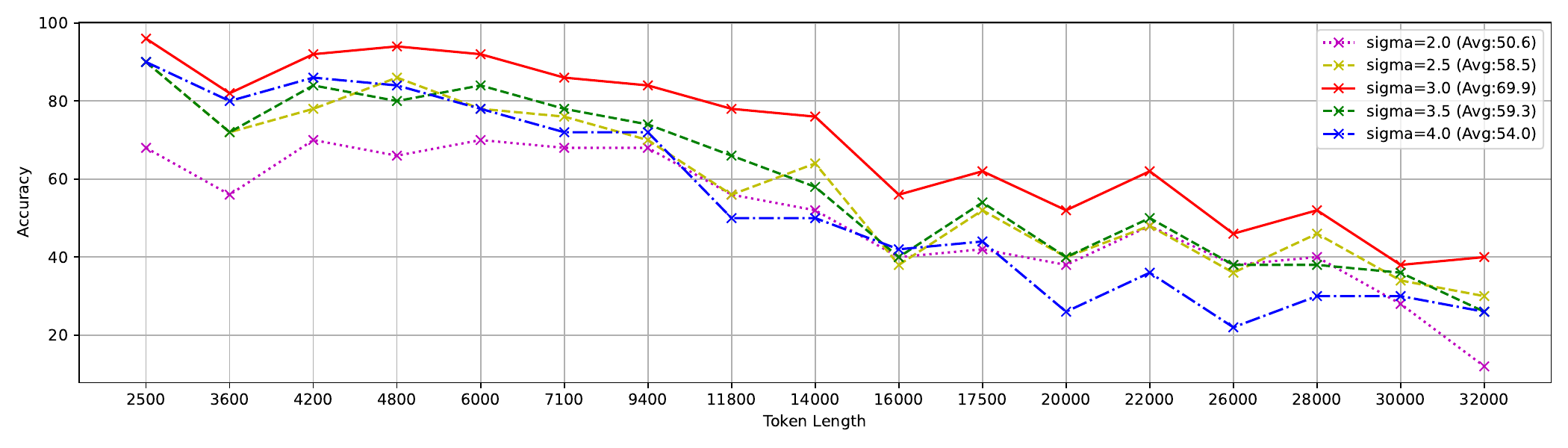}
    \caption{\textbf{Ablation Results (\%) on LongChat-Lines}. Each length consisting of 50 samples. The above are the results of using Linear interpolation on the \texttt{Llama 2-7B} model.}
    \label{fig:sigma}
    \vspace{-2.0em}
\end{figure}

\paragraph{Ablation of Hyperparameters}
In our implementation of truncated Gaussian sampling, as illustrated in \cref{eq:gau_cdf}, the only hyperparameters are the mean $\mu$ and the variance $\sigma$. The mean $\mu$ is determined by the expansion factor. The variance $\sigma$ is adaptable based on data, we conducted experiments with five different values of $\sigma$. The results, as presented in Figure~\ref{fig:sigma}, indicate that the current selection ($\sigma = 3$) yields optimal performance.
% ; for instance, when expanding from $4K$ to $32K$, the expansion factor is $8$, resulting in $mu = \frac{1 + 8}{2}$. 
% but due to the discrete nature of the sampling process, the gradient cannot be back-propagated, leading to a high learning cost. Furthermore, 

\section{Related Works}\label{sec:related_work}
\paragraph{Efficient Transformers and Extra Memory}
FoT~\citep{tworkowski2024focused} addresses the limitations of local attention in transformers by integrating memory attention layers, which enable large models to learn from a wide context while reducing interference. Infini-attention~\citep{munkhdalai2024leave} incorporates compressed memory into the standard attention mechanism and integrates masked local attention and long-term linear attention mechanisms within a single Transformer block. LLoCO~\citep{tan2024lloco} employs LoRA in conjunction with context compression, retrieval, and parameter-efficient fine-tuning to learn context offline. Although these methods can successfully extend the long context window of LLMs, they either require modifications to the attention mechanism or the addition of extra modules for assistance. In contrast, \ours does not require these operations and can be directly applied to a pre-trained model.

\paragraph{Positional Interpolation}
\citet{PI} first proposed extending the context window through positional interpolation, which linearly reduces the input position indices to match the original context window size, thereby preventing catastrophic high attention scores from completely disrupting the self-attention mechanism. Subsequently, various methods (such as NTK~\citep{NTK}, ABF~\citep{ABF}, and EABF~\citep{EABF}) emerged that modify the base frequency of rotary positional encoding to achieve positional interpolation. YaRN~\citep{YaRN} introduced a segmented interpolation method, applying different positional interpolations to different dimensions. LongRoPE~\citep{longrope} identifies and utilizes two forms of non-uniformity in positional interpolation through search, and introduces a progressive expansion strategy for positiona interpolation. Moreover, \ours can be combined with any positional interpolation method.

\paragraph{Positional Encoding}
RandPos~\citep{randpos} first modified position indices so that the model leverages the relativity of positions, enabling it to extend to the target length with fine-tuning over shorter lengths. PoSE~\citep{PoSE} then emphasized the importance of continuous segments, dividing the training length into two parts to further enhance the interpolation effect. \ours utilizes both relativity and continuity, and it also better enables the model to focus on the middle part of the context.

\section{Conclusion}
We proposed \oursfull(\ours), a simple yet effective method to extend the context of large language models. \ours achieves a trade-off between continuity and relativity, enabling the model to exploit positional relativity (\ie, fine-tuning within the pre-trained length), while preserving text continuity (\ie, remaining as close as possible to the pre-trained state). Furthermore, by employing truncated Gaussian sampling, the model can concentrate more on the middle positions during fine-tuning. Experimental results demonstrate that \ours outperforms other methods on both Base and Chat models and effectively mitigates the issue of ``lost in the middle''.

\section*{Acknowledgement}
The authors thank the reviewers for their insightful suggestions to improve the manuscript. This work presented herein is supported by the National Natural Science Foundation of China (62376031).

% \newpage

\medskip

{
\small

\bibliographystyle{unsrtnat}
\bibliography{custom}
}

\newpage

%%%%%%%%%%%%%%%%%%%%%%%%%%%%%%%%%%%%%%%%%%%%%%%%%%%%%%%%%%%%

\appendix

\section{Relative Positional Encoding in RoPE}\label{app:rope}

We provide a simple background proof on the relative positional encoding performed by Rotary Position Embedding~(RoPE)~\cite{su2024roformer}. Given two embedding vectors $\bm{x}_q, \bm{x}_k \in \mathbb{R}^d$ corresponds to query and key at positions $(m, n) \in [0, L)$, where $d$ is embedding dimension, their encoding counterparts can be defined as:
\begin{equation}
\begin{aligned}
\bm{q}_m &= f_q (\bm{x}_q, m) = \mathbf{R}_{\Theta,m}^d (\bm{x}_q, m) \\
\bm{k}_n &= f_k (\bm{x}_k, n) = \mathbf{R}_{\Theta,n}^d (\bm{x}_k, n) \\
\end{aligned}
\end{equation}
where 
\begin{equation}
    \mathbf{R}_{\Theta,m}^d=\begin{bmatrix}
    \cos{m\theta_1} & -\sin{m\theta_1} & \cdots & 0 & 0 \\
    \sin{m\theta_1} & \cos{m\theta_1} & \cdots & 0 & 0 \\
    \vdots & \vdots & \ddots & \vdots & \vdots \\ 
    0 & 0 & \cdots & \cos{m\theta_{d/2}} &  -\sin{m\theta_{d/2}} \\
    0 & 0 & \cdots & \sin{m\theta_{d/2}} &  \cos{m\theta_{d/2}} \\
    \end{bmatrix}
\end{equation}
is the rotary matrix, $\Theta=\{\theta_i=10000^{-2(i-1)/d}, i=[1,2,\dots, d/2]\}$ is pre-defined rotation angles. Then the self attention score can be obtained with:
\begin{equation}
    \begin{aligned}
    \bm{q}_m^{\rm T}\bm{k}_n &= \langle f_q (\bm{x}_q, m), f_k (\bm{x}_k, n)  \rangle \\
    &= {\rm Re} \left[ \sum_{i=0}^{d/2-1} \bm{x}_{q[2i:2i+1]}\bm{x}_{k[2i:2i+1]}^* e^{i(m-n)\theta_i}\right] \\
    &\coloneq g(\bm{x}_m, \bm{x}_n, m - n)
    \end{aligned}
\end{equation}
where $\bm{x}^*$ represents the conjugate complex of $\bm{x}$, $g$ is the derived attention function of RoPE. As seen, RoPE only depends on the relative distances between and encodes the relative position information.

\section{Theoretical findings of \ours design}\label{app:theory}

\begin{theorem}\label{thm:max_relative}
    If $N \ll L$, the spanning size $|D_r|$ of the relative position union in \cref{eq:relative_set} reaches its maximum iff. one of the following groups of inequalities satisfies:\\

    % TODO：补充Proof
    \begin{equation}
     \max(L_h-1, P_e - P_s, L_t-1) + L_h - 1 < P_s < P_e < (L - L_t) / 2,
    \label{eq:ineqgp1}
    \end{equation}
  or
    \begin{equation}
    (L + L_h) / 2 - 1< P_s  < P_e <  L - L_t  - \max(L_h-1, P_e - P_s, L_t-1), \\
    \label{eq:ineqgp2}
    \end{equation} 
where $\max|D_r| = \max(L_h-1, P_e - P_s, L_t-1) + 2N$.
\end{theorem}
\textit{Proof.} Denote four intervals in \cref{eq:relative_set} as $S_i, i = 1, \dots, 4$. According to the inequality of inclusion-exclusion principle for the cardinality of the union of $n$ sets:
\begin{equation}
    |D_r| = |\cup_{i=1}^4 S_i| \leq \sum_{i=1}^4 |S_i|,
\end{equation}
where the equality holds \textit{iff.} all sets are pairwise disjoint. That is
\begin{equation}
    S_i \cap S_j = \varnothing, \quad{} \forall i \neq j
    \label{eq:no_interval}
\end{equation}
Given intervals as in \cref{eq:relative_set}, we have
\begin{equation}
    \begin{aligned}[c]
    \begin{cases}
    {\rm MAX} < P_s - L_h + 1 \\
    P_e < L - L_t - P_e \\
    L - 1 - P_s < L - L_t - L_h + 1
    \end{cases}
    \end{aligned}
    \quad or \quad 
    \begin{aligned}[c]
    \begin{cases}
   {\rm MAX} < L - L_t - P_e \\
    L - 1 - P_s < P_s - L_h + 1 \\
    P_e < L - L_t - L_h + 1 \\
    \end{cases}
    \end{aligned}
    ,
\end{equation}
where ${\rm MAX} =  \max(L_h-1, P_e - P_s, L_t-1)$. The above inequalities can be simplified to \cref{eq:ineqgp1,eq:ineqgp2}.

\begin{lemma}
Under mild assumptions that $L-L_t \approx L$, $L+L_h \approx L$, the maximization in \cref{thm:max_relative} holds for all $(P_s, P_e) \in [N, L/2) \cup (L/2, L-N]$.
\end{lemma}

\textit{Proof.} Given that 
\begin{equation}
\begin{aligned}
    \max(L_h-1, P_e - P_s, L_t-1) + L_h - 1 &< \max(2L_h, N- L_t, N-L_m) < N \\
    L - L_t  - \max(L_h-1, P_e - P_s, L_t-1) &> L - \max(N-L_m, N-L_h, 2L_t) > L - N,
\end{aligned}
\end{equation}
the inequalities in \cref{eq:ineqgp1,eq:ineqgp2} turns into  $[N, L/2) \cup (L/2, L-N]$.

\begin{theorem}
    If $N \ll L$, when the spanning size $|D_r|$  of the relative position union in \cref{eq:relative_set} reaches its maximum, we denote the coverage area of the middle segment as:
    \begin{equation}
        S_m \coloneq \left\{x| x \in [P_s, P_e], (P_s, P_e) \in \left\{\argmax_{(P_s, P_e)} |D_r| \right\}\right\}
    \end{equation}
    thus, we have:
    \begin{equation}
        L \geq S_m + L_h + L_t > L - N/2
    \end{equation}
    
    Furthermore, as $\frac{N}{L} \rightarrow 0$, we have:
    \begin{equation}
        L_h + S_m + L_t \rightarrow L
    \end{equation}
\end{theorem}
% \begin{equation}
% \label{eq:relative_set}
% \small
%     [0, N/3]\cup[P_s-N/3+1, P_e]\cup[L-N/3-P_e, L-1-P_s]\cup[L-2N/3+1, L-1].
% \end{equation}

\section{Experimental Details}
\label{app:exp_details}

\paragraph{Model Hyperparameters}
We fine-tune all models by optimizing the causal language modeling objective.  A learning rate of $2 \times 10^{-5}$ with a linear scheduler is adopted, incorporating 10 warm-up steps. We use the AdamW~\cite{AdamW} optimizer with the hyperparameter configurations specified by PyTorch~\cite{pytorch}. To speed up fine-tuning, we resort to DeepSpeed~\footnote{\url{https://github.com/microsoft/DeepSpeed}} ZeRO stage 1 and Flash Attention-2~\cite{flashattention_v2}. We perform fine-tuning on two A100-80G GPUs with a total batch size of 32 and run inference on a single A100-80G GPU.
For \ours-Base, we fine-tune it for 1,000 steps on a dataset derived from Pile~\cite{The_pile}; for \ours-Chat, we fine-tune it for 100 steps on ShareGPT~\cite{sharegpt}. To ensure fair comparison, we follow the fine-tuning and inference configurations established by~\citet{PoSE}. 

\paragraph{Datasets and Training Cost}
For training the Base model, we directly utilize The Pile data provided by \citet{PoSE}, and select samples with token lengths exceeding 4K. For training the Chat model, we filter the ShareGPT data from public datasets\footnote{\url{https://huggingface.co/datasets/Aeala/ShareGPT_Vicuna_unfiltered}}. Specifically, we used the Vicuna prompt template to sequentially concatenate the ShareGPT data until each data point comprises at least 4K tokens. Then, we select 3.2K data points to train for 100 steps. Particularly, during the instruction tuning process, we mask the USER part and allow the model to calculate the loss only on the ASSISTANT part. We utilize two A100-80G machines with a global batch size of 32, fully utilizing the available memory. Running 1,000 steps for the Base model takes approximately 6 hours, while running 100 steps for the Chat model takes approximately 2 hours.

\section{Robustness Across LLMs}

Our proposed method exhibits strong generalization capabilities and can be applied to other large language models (LLMs) without the need for parameter modification. To validate this, we conducted experiments on \texttt{Baichuan2-7B}, with the corresponding results presented in Table~\ref{tab:baichuan2}. 

\begin{table}[htbp]
    \centering
    \small
    \setlength\tabcolsep{12.9pt} % 列间距
    \caption{\textbf{Perplexity results of GovReport and Proof-pile.} Each experiment is the average perplexity of 50 samples, and all results are based on \texttt{Baichuan2-7B} fine-tuned on 4K data length.}
    \resizebox{\linewidth}{!}{
    \begin{tabular}{@{}lcccccccccc@{}}
    \toprule
    % & \multirow{3}{*}{\begin{tabular}[c]{@{}c@{}}\textbf{Context Size}\\ \textbf{Train / Target}\end{tabular}}
    \multirow{3}{*}{\textbf{Model}} & \multicolumn{4}{c}{\textbf{GovReport}} & \multicolumn{4}{c}{\textbf{Proof-pile}} \\
    \cmidrule(lr){2-5} \cmidrule(lr){6-9}
    & \textbf{4K} & \textbf{8K} & \textbf{16K} & \textbf{32K} & \textbf{4K} & \textbf{8K} & \textbf{16K} & \textbf{32K} \\ \midrule
    Original & 3.3 & - & - & - & 5.8 & - & - & - \\
    \ours-Linear & 3.6 & 2.9 & 2.5 & 2.2 & 6.2 & 6.1 & 6.0 & 5.8 \\
    \bottomrule
    \label{tab:baichuan2}
    \end{tabular}
    }
\end{table}

Furthermore, we fine-tuned \texttt{LLaMa3-8B} using a context window size of $4K$ tokens, with the experimental outcomes shown in Table~\ref{tab:llama3}.

\begin{table}[htbp]
    \centering
    \small
    \caption{\textbf{Results (\%) on LongChat-Lines}. Each length consists of 50 samples. All results are fine-tuned on \texttt{Llama-3-8B} with \textbf{4K} length data through linear position interpolation.}
    \resizebox{\linewidth}{!}{
    \begin{tabular}{@{}lccccccccccccccccc@{}}
    \toprule
   \textbf{AVG Length} & \textbf{2000} & \textbf{4000} & \textbf{7800} & \textbf{8800} & \textbf{9700} & \textbf{11000} & \textbf{12000} & \textbf{14000} & \textbf{17000} & \textbf{19000} & \textbf{24000} & \textbf{28000} & \textbf{32000} \\ \midrule
    \ours-Linear & 0.98 & 1.00 & 0.96 & 0.94 & 0.86 & 0.92 & 0.92 & 0.92 & 0.86 & 0.84 & 0.70 & 0.60 & 0.48 \\
    \bottomrule
    \label{tab:llama3}
    \end{tabular}
    }
\end{table}

The results in Tables \ref{tab:baichuan2} and \ref{tab:llama3} clearly demonstrate the transferability of our method to different models, underscoring its robustness. Of particular note is that despite \texttt{LLaMa3-8B} having a native context length of $8K$ tokens, fine-tuning on training data with a $4K$ context window yielded unexpectedly strong performance.

\section{LongChat Lines Results}
\label{app:longchat_lines}

\begin{figure}[htbp]
    \centering
    \includegraphics[width=1.0\textwidth]{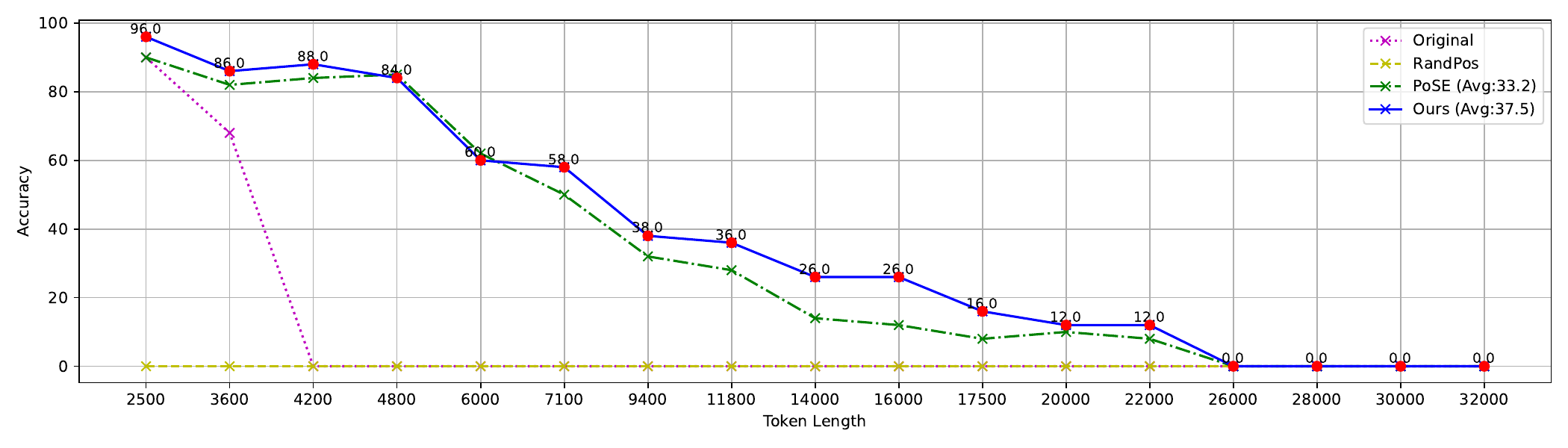}
    \caption{\textbf{Results (\%) on LongChat-Lines}. Each length consisting of 50 samples. The above are the results of using NTK interpolation on the \texttt{Llama 2-7B} model.}
    \label{fig:longchat_lines_ntk}
\end{figure}

\begin{figure}[htbp]
    \centering
    \includegraphics[width=1.0\textwidth]{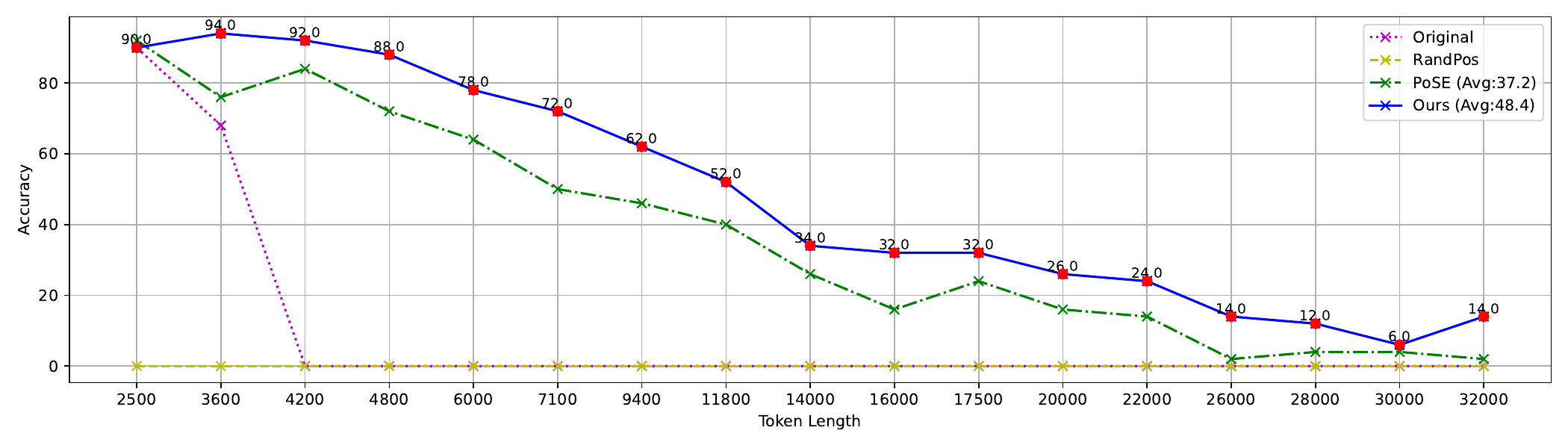}
    \caption{\textbf{Results (\%) on LongChat-Lines}. Each length consisting of 50 samples. The above are the results of using Yarn interpolation on the \texttt{Llama 2-7B} model.}
    \label{fig:longchat_lines_yarn}
\end{figure}

The interpolation methods using NTK and Yarn are presented in Figures \ref{fig:longchat_lines_ntk} and \ref{fig:longchat_lines_yarn}. As can be seen, \ours performs the same as the Linear method for interpolation, still outperforming other methods. The result of NTK at 26K-32K is zero, which is due to the inherent properties of NTK, a finding that is aligns with \citet{PoSE}.


\section{LongBench Subtasks Results}
\label{app:longbench_subtasks}

The results of each subtask in Tables~\ref{tab:longbench2} are shown in Tables ~\ref{tab:subtask1} and \ref{tab:subtask2}.

\begin{table}[htbp]
\centering
% \small
% \setlength\tabcolsep{2pt} % 列间距
\caption{\textbf{Experimental results (\%) of the \texttt{LongBench} subtasks} selected in \citet{EABF}. $^\dag$ indicates results quoted from \citet{EABF}. \textbf{Len} represents the context length during fine-tuning. All results are based on \texttt{Llama 2-7B}.}
\resizebox{\linewidth}{!}{
\begin{tabular}{@{}lcccccccccccccccc@{}}
\toprule
\multirow{3}{*}{\textbf{Model}} & \multirow{3}{*}{\textbf{Num / Len}} & \multicolumn{3}{c}{\textbf{Singl-Doc QA}} & \multicolumn{3}{c}{\textbf{Multi-Doc QA}} & \multicolumn{3}{c}{\textbf{Summarization}} & \multicolumn{3}{c}{\textbf{Few-shot Learning}} & \multirow{3}{*}{\textbf{AVG}} \\ 
\cmidrule(r){3-5} \cmidrule(lr){6-8} \cmidrule(lr){9-11} \cmidrule(l){12-14}
 & & \textbf{NQA} & \textbf{QAPR} & \textbf{MFQA\_en} & \textbf{HPQA} & \textbf{WMQA} & \textbf{MSQ} & \textbf{GR} & \textbf{QMSM} & \textbf{MNWS} & \textbf{TREC} & \textbf{TRVQA} & \textbf{SMSM} &  &  \\
\midrule
% Llama2-7B-chat-4k$^*$   & - & 18.7 & 19.2 & 36.8 & 25.4 & 32.8 & 9.4 & 27.3 & 20.8 & 25.8 & 61.5 & 77.8 & 40.7 & 33.0     \\
% LongChat-v1.5-7B-32k$^*$    & - & 16.9 & 27.7 & 41.4 & 31.5 & 20.6 & 9.7 & 30.8 & 22.7 & 26.4 & 63.5 & 82.3 & 34.2 & 34.0     \\
% Vicuna-v1.5-7B-16k$^*$      & - & 19.4 & 26.1 & 38.5 & 25.3 & 20.8 & 9.8 & 27.9 & 22.8 & 27.2 & 71.5 & 86.2 & 40.8 & 34.7     \\ \midrule
PI$^\dag$                      & \multirow{5}{*}{3.5K / 16K} & 20.1 & 30.4 & 45.3 & 26.1 & 30.1 & 9.9 & 28.1 & 23.7 & 26.6 & 68.0 & 84.9 & 42.5 & 36.3     \\
NTK-By-Parts$^\dag$            &  & 15.9 & 31.1 & 40.1 & 25.4 & 26.6 & 7.2 & 26.7 & 22.4 & 26.9 & 68.5 & 82.8 & 42.9 & 34.7     \\
Yarn$^\dag$                    &  & 20.3 & 28.9 & 42.8 & 27.8 & 30.7 & 7.2 & 27.4 & 22.5 & 26.8 & 66.0 & 85.6 & 42.6 & 35.7     \\
ABF$^\dag$                     &  & 24.6 & 32.8 & 45.6 & 35.1 & 30.3 & 15.2 & 30.8 & 23.0 & 27.4 & 71.0 & 84.7 & 42.7 & 38.6     \\
EABF$^\dag$     &  & 21.9 & 31.0 & 47.1 & 40.1 & 32.7 & 15.1 & 32.3 & 23.0 & 27.1 & 70.5 & 86.7 & 42.0 & 39.1     \\ \midrule
\ours{}                  & \cellcolor[HTML]{FFFC9E}{\textbf{3.2}K / \textbf{4}K} & 23.0 & 34.6 & 46.8 & 42.2 & 33.7 & 17.4 & 30.4 & 24.3 & 26.8 & 69.5 & 84.0 & 41.9 & \cellcolor[HTML]{FFFC9E}{\textbf{39.6}}     \\
\bottomrule
\label{tab:longbench1}
\end{tabular}
}
\end{table}

\begin{table}[htbp]
\centering
\small
\caption{\textbf{Experimental results (\%) of the LongBench subtasks.}}
\resizebox{\linewidth}{!}{
\begin{tabular}{@{}lcccccccccc@{}}
\toprule
\multirow{3}{*}{\textbf{Model}} & \multicolumn{3}{c}{\textbf{Singl-Doc QA}} & \multicolumn{3}{c}{\textbf{Multi-Doc QA}} & \multicolumn{3}{c}{\textbf{Summarization}} \\ 
\cmidrule(r){2-4} \cmidrule(lr){5-7} \cmidrule(lr){8-10} & \textbf{NQA} & \textbf{QAPR} & \textbf{MFQA\_en} & \textbf{HPQA} & \textbf{WMQA} & \textbf{MSQ} & \textbf{GR} & \textbf{QMSM}  & \textbf{MNWS} \\
\midrule  
Llama2-7B-chat-4k$^*$     & 18.7 & 19.2 & 36.8 & 25.4 & 32.8 & 9.4  & 27.3 & 20.8 & 25.8 \\
XGen-7B-8k$^*$            & 18.0 & 18.1 & 37.7 & 29.7 & 21.1 & 10.3 & 27.3 & 20.5 & 26.2 \\
InternLM-7B-8k$^*$        & 12.1 & 16.7 & 23.4 & 28.7 & 22.8 & 9.0  & 9.7  & 15.9 & 22.8 \\
Vicuna-v1.5-7B-16k$^*$    & 19.4 & 26.1 & 38.5 & 25.3 & 20.8 & 9.8  & 27.9 & 22.8 & 27.2 \\
LongChat-v1.5-7B-32k$^*$  & 16.9 & 27.7 & 41.4 & 31.5 & 20.6 & 9.7  & 30.8 & 22.7 & 26.4 \\ \midrule
\ours                 & 23.0 & 34.6 & 46.8 & 42.2 & 33.7 & 17.4 & 30.4 & 24.3 & 26.8  \\
\bottomrule
\label{tab:subtask1}
\end{tabular}
}
\end{table}

\begin{table}[htbp]
\centering
\small
\setlength\tabcolsep{8pt} % 列间距
\caption{\textbf{Experimental results (\%) of the LongBench subtasks.}}
% \resizebox{\linewidth}{!}{
\begin{tabular}{@{}lccccccc@{}}
\toprule
\multirow{3}{*}{\textbf{Model}} & \multicolumn{3}{c}{\textbf{Few-shot Learning}} & \multicolumn{2}{c}{\textbf{Code Completion}} & \multicolumn{2}{c}{\textbf{Synthetic Tasks}} \\ 
\cmidrule(r){2-4} \cmidrule(lr){5-6} \cmidrule(lr){7-8} & \textbf{TREC} & \textbf{TRVQA} & \textbf{SMSM} & \textbf{PC} & \textbf{PR\_en} & \textbf{LCC} & \textbf{RBP} \\
\midrule  
Llama2-7B-chat-4k$^*$     & 61.5 & 77.8 & 40.7 & 2.1 & 9.8 & 52.4 & 43.8 \\
XGen-7B-8k$^*$            & 65.5 & 77.8 & 25.3 & 2.1 & 8.5 & 38.6 & 38.6 \\
InternLM-7B-8k$^*$        & 52.0 & 77.8 & 21.2 & 3.0 & 6.0 & 44.1 & 28.8 \\
Vicuna-v1.5-7B-16k$^*$    & 71.5 & 86.2 & 40.8 & 6.5 & 4.5 & 51.0 & 43.5 \\
LongChat-v1.5-7B-32k$^*$  & 63.5 & 82.3 & 34.2 & 1.0 & 30.5 & 53.0 & 55.3 \\ \midrule
\ours                 & 69.5 & 84.0 & 41.9 & 3.0 & 11.0 & 52.0 & 48.7  \\
\bottomrule
\label{tab:subtask2}
\end{tabular}
% }
\end{table}

It is noteworthy that, to provide further evidence of the efficacy of our model, we have specifically chosen 12 tasks from the four categories outlined in \citet{EABF} for comparison purposes. As delineated in Table \ref{tab:longbench1}, we are able to attain superior performance on LongBench in comparison to EABF~\cite{EABF}, even with shorter training lengths and less data.

\section{Limitations}

When extending the context beyond the pre-trained length, there is an inevitable loss of information due to position interpolation, particularly when fine-tuning is restricted to the pre-trained length. However, in comparison to previous methods such as RandPos~\cite{randpos} and PoSE~\cite{PoSE}, \ours has effectively mitigated the issue of ``Lost-in-the-Middle'' by introducing truncated Gaussian sampling. Additionally, as discussed in reference ~\citet{Lost_in_the_Middle}, decoder-only models are prone to inherently exhibiting a U-shaped performance curve on this task. Therefore, completely solving this problem remains challenging.

% \section{Modified Yarn}

\section{Loss Curve}

\begin{figure}[h]
    \centering
    \begin{subfigure}{0.48\textwidth}
        \centering
        \includegraphics[width=\linewidth]{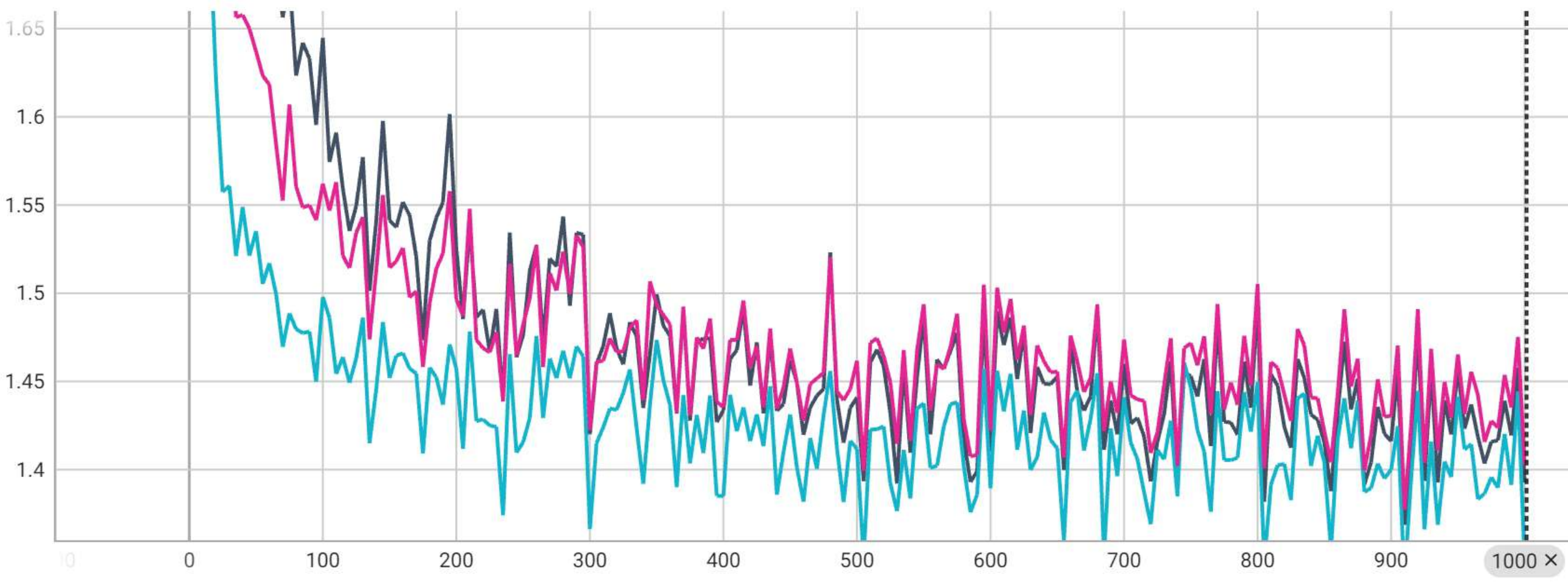}
        \caption{\ours-192K Training Loss}
    \end{subfigure}
    \hfill
    \begin{subfigure}{0.48\textwidth}
        \centering
        \includegraphics[width=\linewidth]{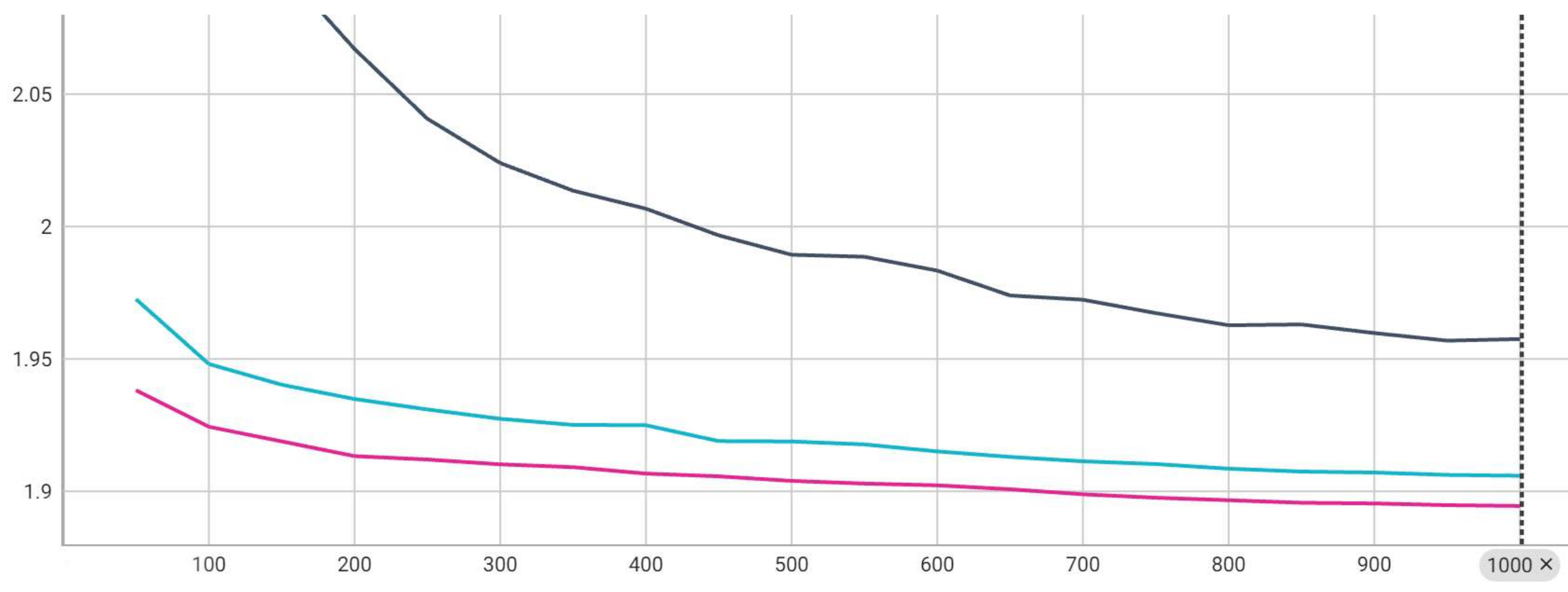}
        \caption{\ours-192K Validation Loss}
    \end{subfigure}
    \begin{subfigure}{0.48\textwidth}
        \centering
        \includegraphics[width=\linewidth]{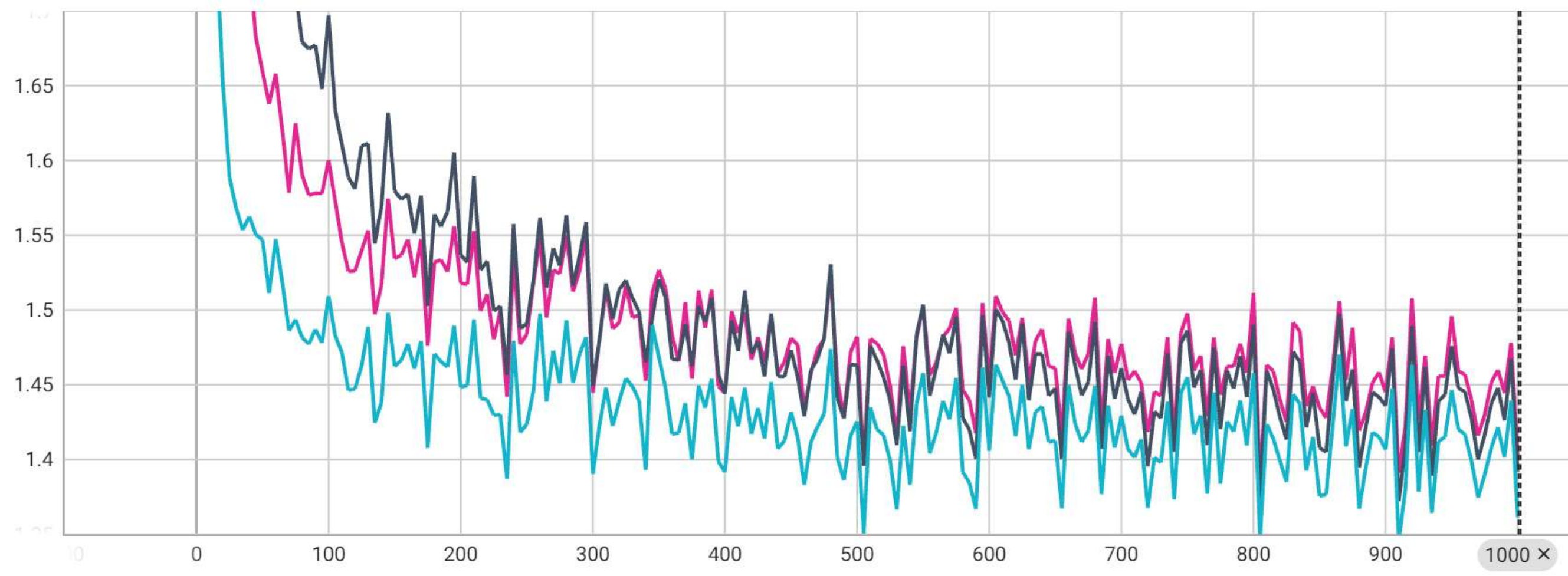}
        \caption{\ours-256K Training Loss}
    \end{subfigure}
    \hfill
    \begin{subfigure}{0.48\textwidth}
        \centering
        \includegraphics[width=\linewidth]{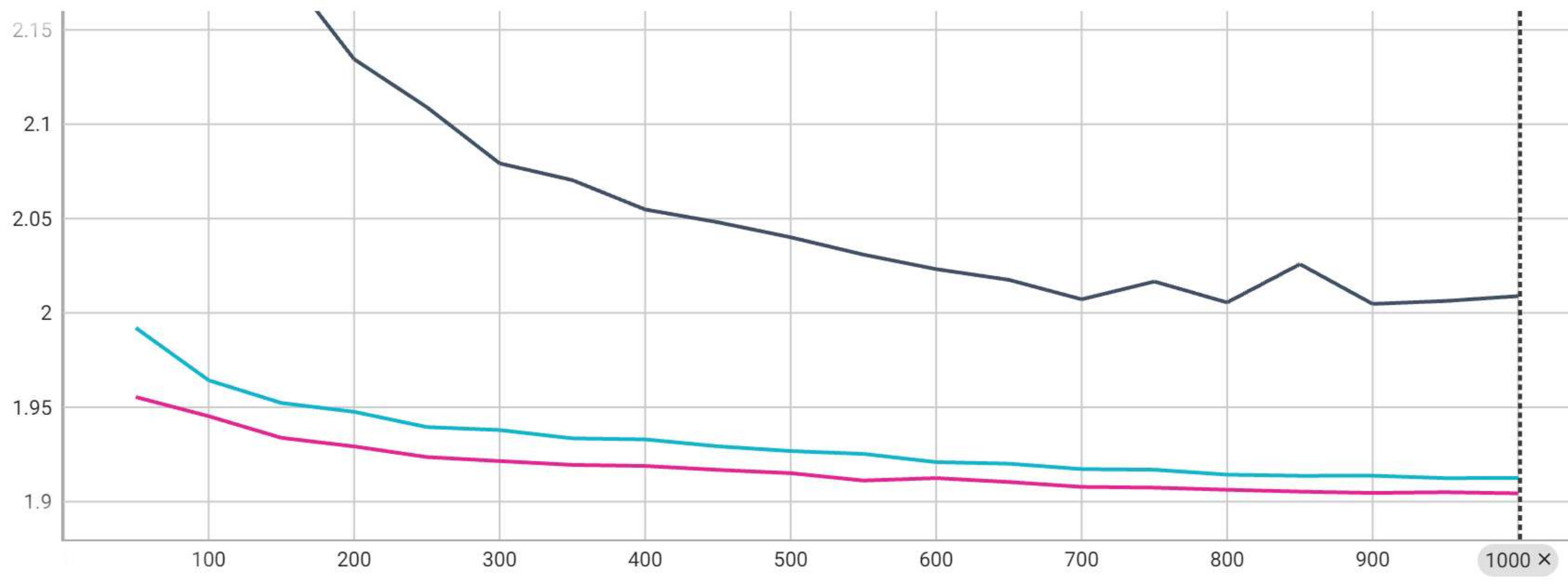}
        \caption{\ours-256K Validation Loss}
    \end{subfigure}
    % \hfill
    \caption{\textbf{Fine-tuning loss curve based on \texttt{Llama 2-7B}.}The black line represents Linear interpolation, the pink line represents NTK interpolation, and the cyan line represents YaRN interpolation.}
    \label{fig:loss_curve}
\end{figure}

\begin{figure}[h]
    \centering
    \begin{subfigure}{0.48\textwidth}
        \centering
        \includegraphics[width=\linewidth]{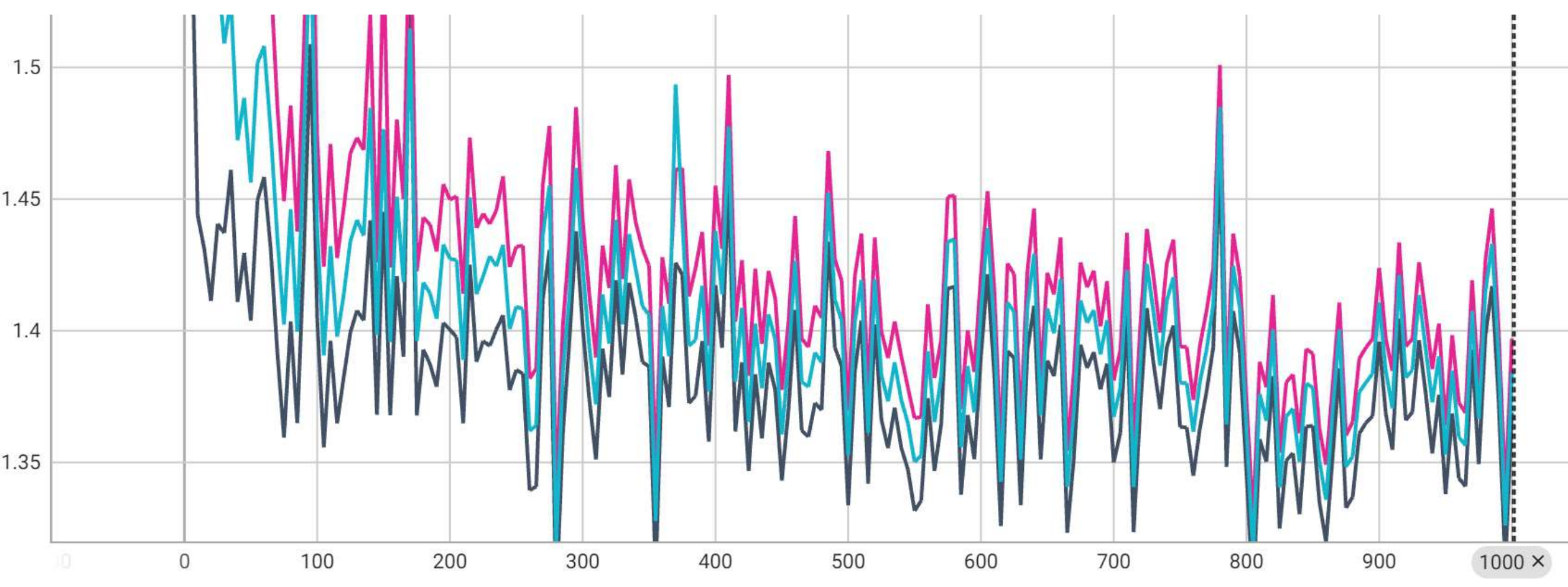}
        \caption{RandPos Training Loss}
    \end{subfigure}
    \hfill
    \begin{subfigure}{0.48\textwidth}
        \centering
        \includegraphics[width=\linewidth]{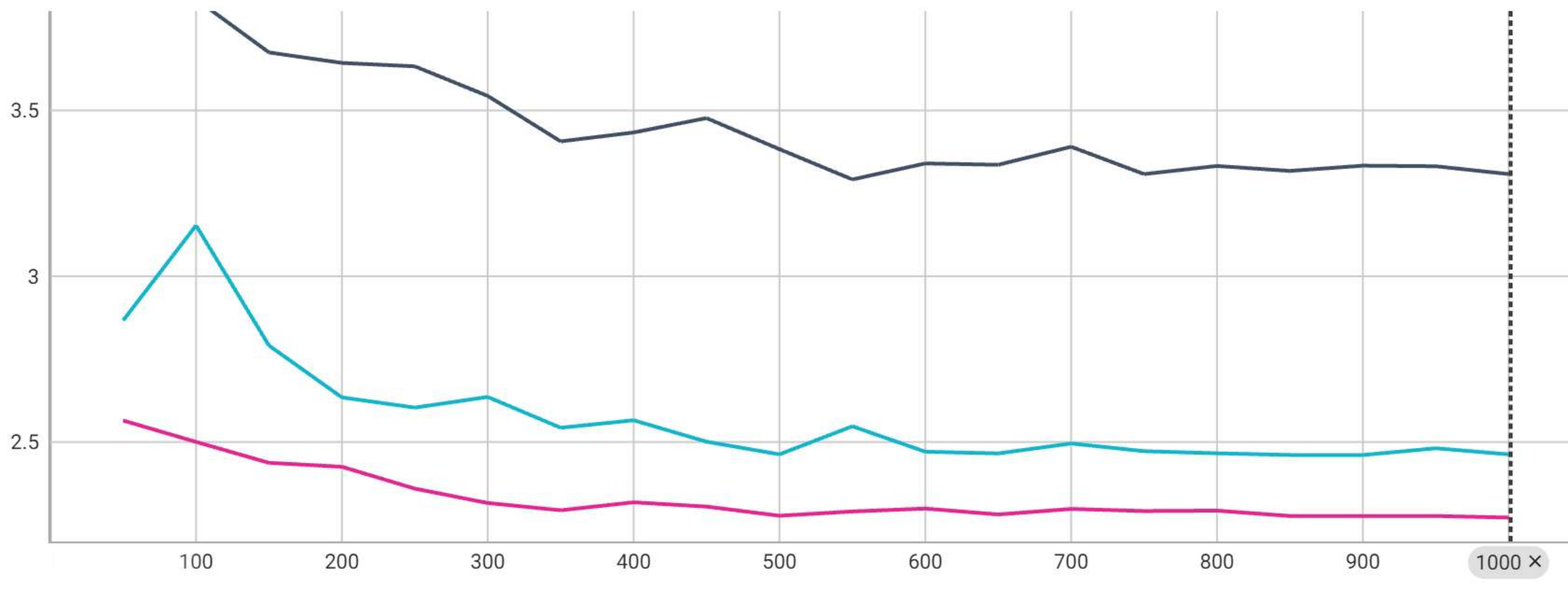}
        \caption{RandPos Validation Loss}
    \end{subfigure}
    \begin{subfigure}{0.48\textwidth}
        \centering
        \includegraphics[width=\linewidth]{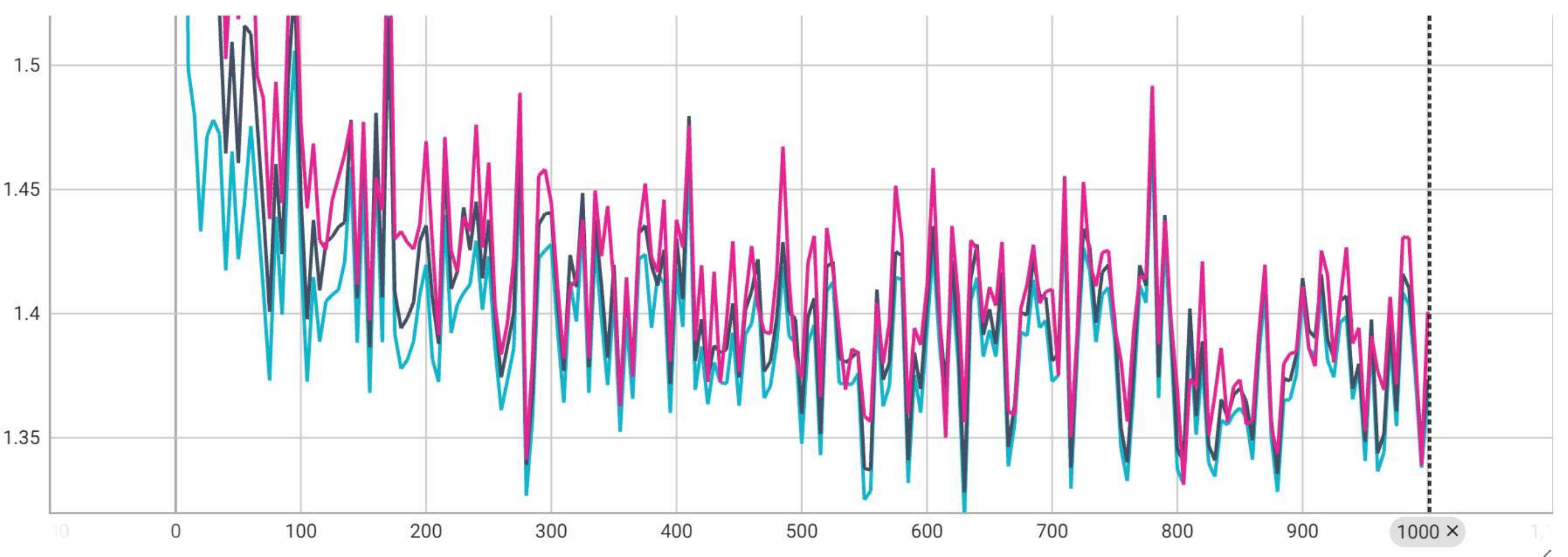}
        \caption{PoSE Training Loss}
    \end{subfigure}
    \hfill
    \begin{subfigure}{0.48\textwidth}
        \centering
        \includegraphics[width=\linewidth]{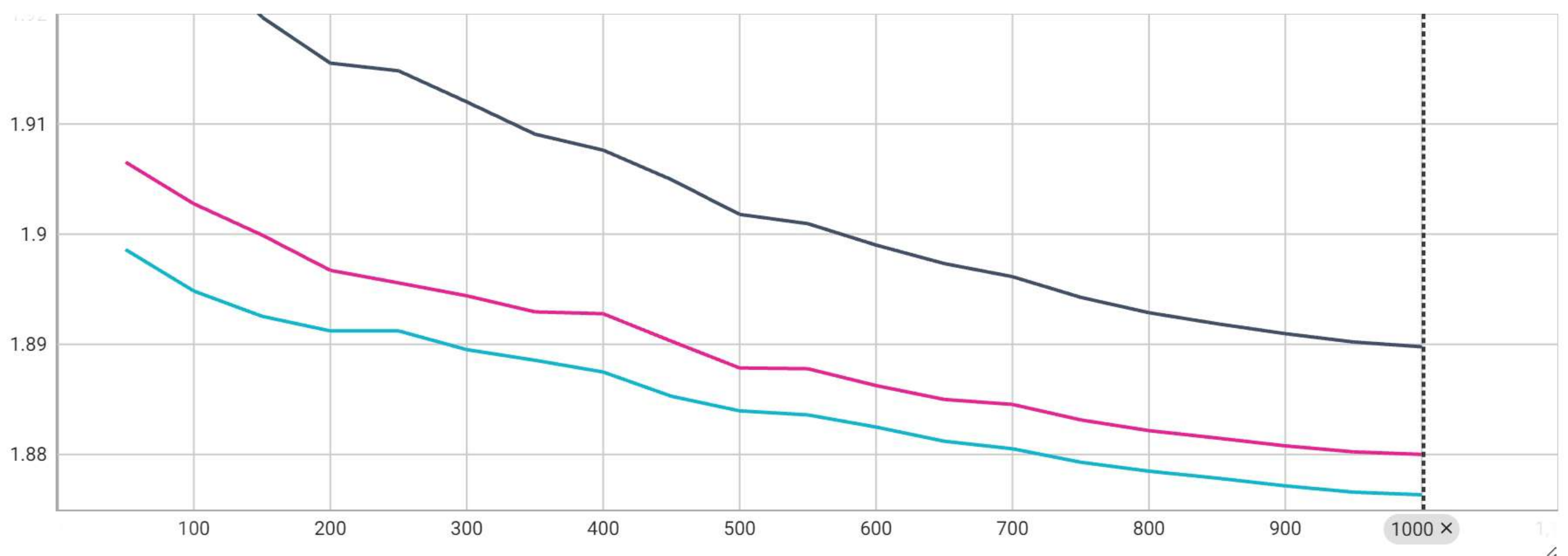}
        \caption{PoSE Validation Loss}
    \end{subfigure}
    \begin{subfigure}{0.48\textwidth}
        \centering
        \includegraphics[width=\linewidth]{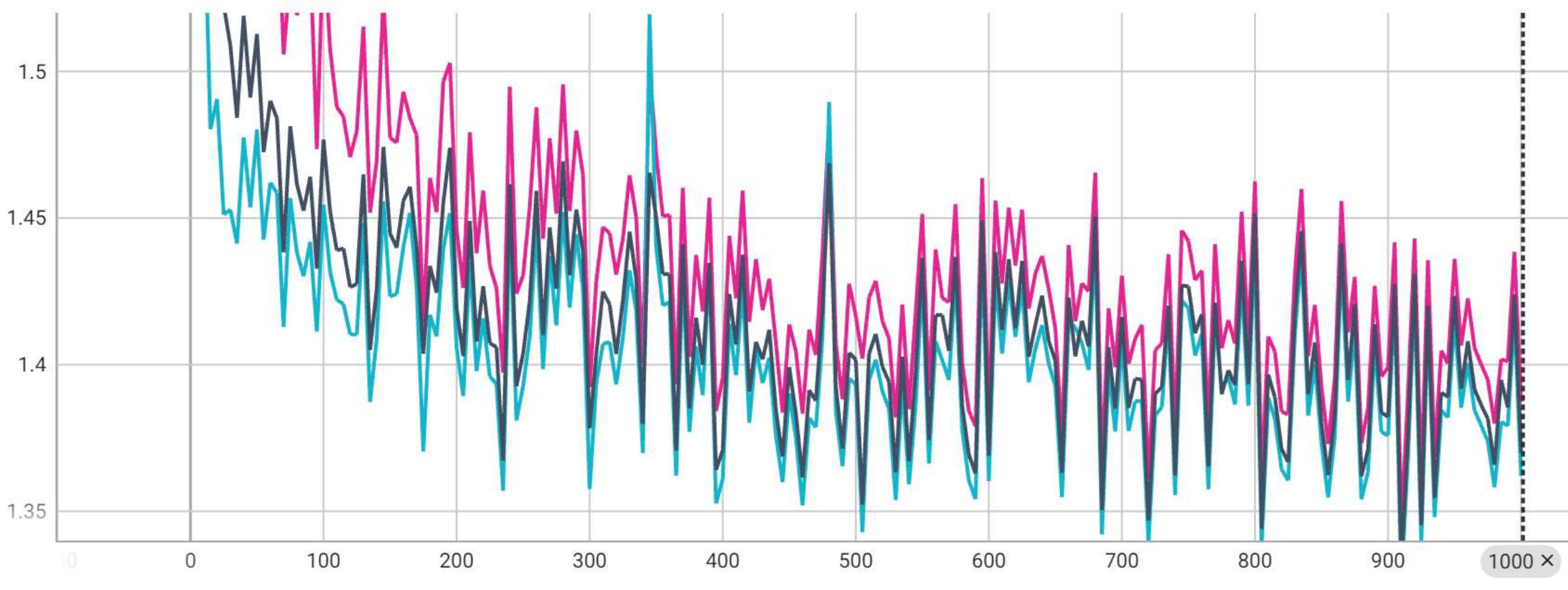}
        \caption{\ours Training Loss}
    \end{subfigure}
    \hfill
    \begin{subfigure}{0.48\textwidth}
        \centering
        \includegraphics[width=\linewidth]{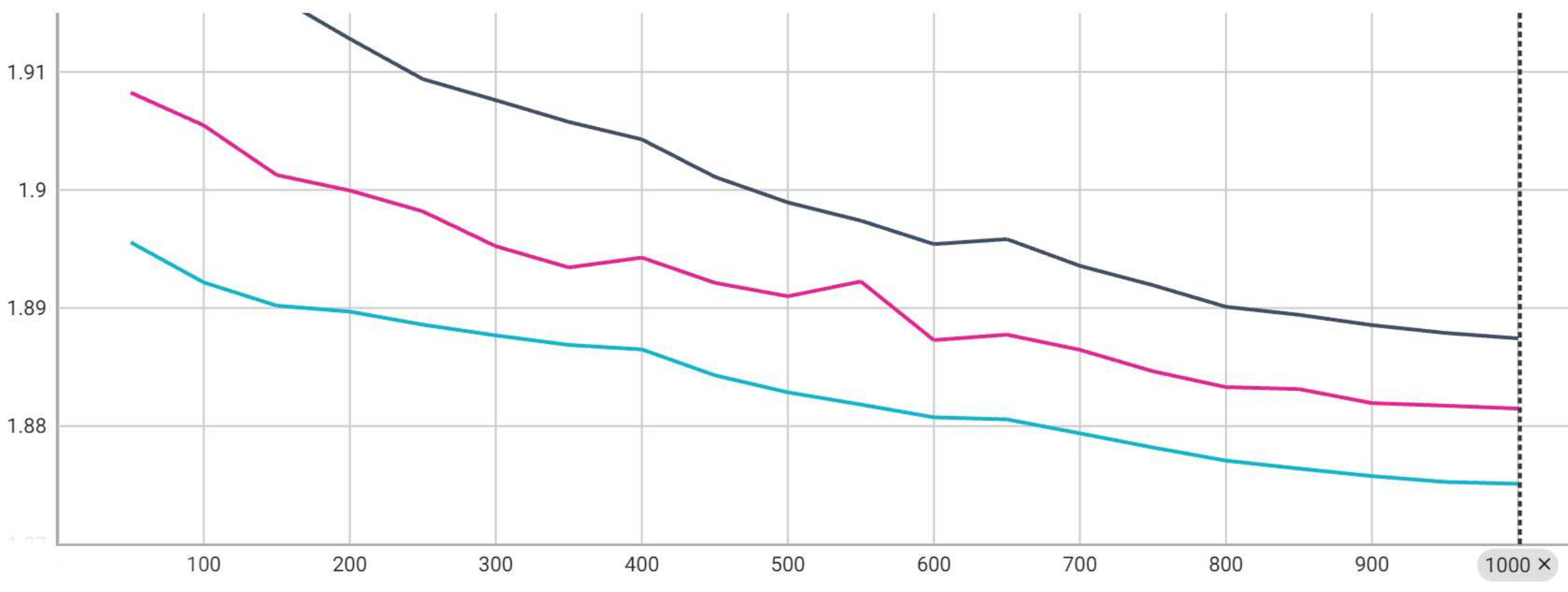}
        \caption{\ours Validation Loss}
    \end{subfigure}
    \hfill
    \caption{\textbf{Fine-tuning loss curve based on \texttt{Llama 2-7B}.} The black line represents Linear interpolation, the pink line represents NTK interpolation, and the cyan line represents YaRN interpolation.}
    \label{fig:needle}
\end{figure}

\end{document}